\documentclass[acmsmall]{acmart}

\usepackage{multirow}

\AtBeginDocument{%
  \providecommand\BibTeX{{%
    \normalfont B\kern-0.5em{\scshape i\kern-0.25em b}\kern-0.8em\TeX}}}

\setcopyright{acmcopyright}
\copyrightyear{2021}
\acmYear{2021}
\acmDOI{10.1145/XXXX.XXXX}
\acmJournal{CSUR}
\acmPrice{15.00}
\acmISBN{978-1-4503-XXXX-X/18/06}

\begin{document}

\title{A Survey on Spatio-temporal Data Analytics Systems}

\author{Md Mahbub Alam}
\email{mahbub.alam@dal.ca}
\author{Luis Torgo}
\email{ltorgo@dal.ca}
\affiliation{%
  \institution{Dalhousie University}
  \streetaddress{6299 South St.}
  \city{Halifax}
  \state{NS}
  \country{Canada}
  \postcode{B3H 4R2}
}

\author{Albert Bifet}
\affiliation{%
  \institution{The University of Waikato}
  \streetaddress{Knighton Road}
  \city{Hamilton}
  \country{New Zealand}}
\email{abifet@waikato.ac.nz}

\renewcommand{\shortauthors}{Alam, et al.}

\begin{abstract}
Due to the surge of spatio-temporal data volume, the popularity of location-based services and applications, and the importance of extracted knowledge from spatio-temporal data to solve a wide range of real-world problems, a plethora of research and development work has been done in the area of spatial and spatio-temporal data analytics in the past decade. The main goal of existing works was to develop algorithms and technologies to capture, store, manage, analyze, and visualize spatial or spatio-temporal data. The researchers have contributed either by adding spatio-temporal support with existing systems, by developing a new system from scratch for processing spatio-temporal data, or by implementing algorithms for mining spatio-temporal data. The existing ecosystem of spatial and spatio-temporal data analytics can be categorized into three groups, (1) spatial databases (SQL and NoSQL), (2) big spatio-temporal data processing infrastructures, and (3) programming languages and software tools for processing spatio-temporal data. Since existing surveys mostly investigated big data infrastructures for processing spatial data, this survey has explored the whole ecosystem of spatial and spatio-temporal analytics along with an up-to-date review of big spatial data processing systems. This survey also portrays the importance and future of spatial and spatio-temporal data analytics. 

\end{abstract}



\maketitle

\section{Introduction}
\label{sec:intro}
Due to the technological advancement of the internet, sensor devices, GPS-enabled devices and the popularity of location-based services (LBS) and applications (such as map services, recommendation systems, navigation systems, location-based social networks, and other applications), a huge volume of geo-referenced data is generated every day, often called big spatial data. However, a significant portion of these data comes with a timestamp (temporal-tag), leading to spatio-temporal data. As a result of the availability of smart mobile devices and the internet, LBS applications and services are part of our daily activities and contributing significantly to this data growth. These data also comes from other sources such as vehicles, sensors positioned around the world, satellites, space telescopes, aerial photography, land survey, medical imaging and more. Therefore,  mining information from this huge volume of spatio-temporal data is not only important for popular LBS applications and services we are using today, but it is also important for scientific discovery and exploration of a wide range of application domains, such as climate change analysis, earthquake analysis, weather forecasting, urban planning, health care, modern transportation system, agriculture, space exploration, crime data analysis, e-commerce and advertising, epidemic analysis, animal migration, oceanography, and many more. In this context, there is a demand for efficient tools and data processing systems to store, manage, analyze, and visualize the high dimensional and heterogeneous big spatio-temporal data.

Research and development of spatial and spatio-temporal database systems have started with traditional relational database systems (RDBMSs). Traditional RDBMSs with a spatial extension (such as PostgreSQL/PostGIS~\cite{url:PostGIS}, Oracle Spatial, and more) are efficient in a single node computing environment. However, due to the lack of parallelism and the I/O bottleneck, these systems only work well for  relatively small datasets. Besides, these systems have limited analysis and visualization capabilities. Therefore, one may question if spatial RDBMSs are significant in this era of big spatial data. Researchers are continuously adding new features to make these systems adaptable in this new era. Researchers have also developed a few parallel and distributed systems by using spatial RDBMSs. Though current spatial RDBMSs are not massively scalable, these systems are scalable enough to solve many real-world problems we are facing today. There is a huge demand for Spatial RDBMSs in a wide range of application domains at the enterprise level. Therefore, spatial RDBMSs are still significant in this era of big spatial and spatio-temporal data.


Along with the limitations mentioned above, traditional RDBMSs also did not have support to store and process semi-structured or unstructured data. Therefore, NoSQL (Not-Only-SQL) database systems (e.g., MongoDB, Cassandra)~\cite{Felix2017_NoSQL_Survey, Ali2018_NoSQL_Survey} have emerged as alternative databases, which are schemaless, highly available, and horizontally scalable. Currently, a few of these systems have limited native support to store and process spatial data. Still, researchers have extended some of these databases to add spatial support. Moreover, several big spatial data processing systems have been developed by utilizing the power of NoSQL databases~\cite{Nishimura2011_MD-HBase, Hughes2015_GeoMesa, Qin2019_THBase, Li2020_JUST, Li2020_TrajMesa}. Spatial support of current NoSQL databases lack spatial analysis and visualization, and only a few of them support SQL-like query language. Although many of the present modern RDBMSs do have support for processing data other than structured data. For example, GeoJSON support in PostgreSQL/PostGIS.

In recent years, a number of data processing systems have  emerged to process big spatial and spatio-temporal data. These systems are implemented mainly by extending the MapReduce framework Hadoop~\cite{url:hadoop}, Spark~\cite{url:spark, Zaharia2010_Spark_1, Zaharia2016_Spark_2}, and NoSQL database systems~\cite{Felix2017_NoSQL_Survey, Ali2018_NoSQL_Survey} to incorporate spatial and temporal data types, partitioning and indexing techniques, geometric operations, and a SQL-like query language. However, a few of them have been developed either from scratch~\cite{Becla2013_SciDB, Baumann1997_RasDaMan, Maria2019_DISTIL} or by extending systems other than Hadoop, Spark, and NoSQL databases~\cite{Eldawy2017_Sphinx, Alamoudi2015_AsterixDB, Mahbub2018_BigSpatial_PStudy}. Recently, Python libraries such as DASK~\cite{Team2016_Dask} and RAPIDS~\cite{2018_RAPIDS} emerged as parallel and distributed platforms for processing big data. These big data systems can be either spatial~\cite{Eldawy2015_SpatialHadoop, Yu2015_GeoSpark1, Nidzwetzki2018_BBoxDB}, spatio-temporal~\cite{Alarabi2018_ST-Hadoop, Stefan2017_STARK}, trajectory~\cite{Bakil2019_HadoopTrajectory, Zhang2017_TrajSpark, Li2020_JUST}, or spatial stream~\cite{Mahmood2017_Tornado, Chen2020_SSTD, Shaikh2020_GeoFlink} data processing systems. However, these big data systems have limited analysis and visualization capabilities. Also, a few of them support SQL-like query language, but not as efficiently as spatial RDBMSs.

Due to the heterogeneity and implicit spatial and temporal dependencies of spatio-temporal data, being able to extract and analyze knowledge from these data can be extremely challenging. The data mining, analysis, and visualization support of existing spatial RDBMSs, spatial NoSQL databases, and big spatial data infrastructures, are very limited. There is a wide range of libraries and packages  available for mining, analyzing, and visualizing spatial, spatio-temporal, and trajectory data in two popular de facto programming languages for data science, R~\cite{Team2020_R} and Python~\cite{van1995_python}. Besides, a large community of people is continuously working to introduce new libraries and packages to meet the current and future demands. However, these libraries and packages can not store and process a large volume of data. Therefore, the utilization of these libraries and packages with RDBMSs, NoSQL databases, and big data infrastructures in essential to fill the gap. On the other hand, GIS software like ArcGIS~\cite{url:ArcGIS} and QGIS~\cite{QGIS_software} are leading tools to collect, store, process, and visualize spatial data. Initially, GIS software has developed for a single user with limited DBMS capability. Currently, GIS software like ArcGIS has developed additional tools to utilize the processing capability of Hadoop and Spark. ArcGIS can also process data stored in spatial RDBMSs. Moreover, Python and R users can use the functionality of GIS software to process spatial, spatio-temporal, and trajectory data.

\begin{figure}[!htbp]
	\centering
	\includegraphics[width=0.65\linewidth, keepaspectratio]{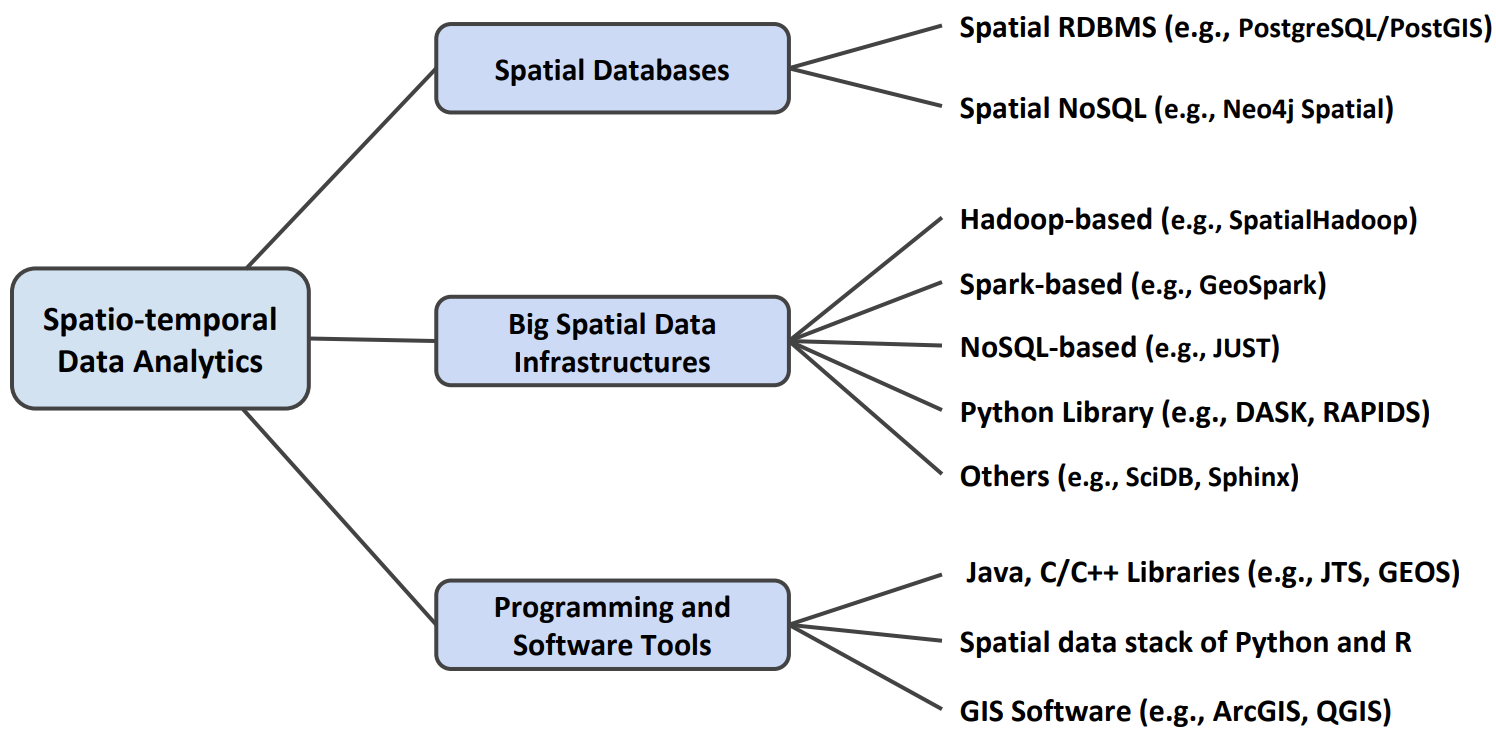}
	\caption{The Ecosystem of Spatio-temporal Data Analytics}
	\label{fig:st_ecosystem}
\end{figure}

Researchers from both academia and industry are working in order to meet the current and future demands for big spatio-temporal analytics. Surveys are always crucial for current and future researchers to know about the state-of-the-art. Surveys also help to choose a data analytics system based on application requirements. However, existing surveys and performance analyses are not up-to-date and mostly summarize infrastructures related to big spatial data processing. In the meantime, a number of big spatio-temporal, trajectory, and spatial stream data processing systems have emerged along with a few new spatial data processing systems. Besides, existing surveys have not considered spatial RDBMSs, GIS software, and spatial support in programming languages. These shortcomings motivate the current paper. This survey categorizes the existing ecosystem of spatio-temporal data analytics (see Figure~\ref{fig:st_ecosystem}) into three data dimensional categories: (1) Data Storage: spatial RDBMSs and NoSQL databases, by first defining the significance of spatial RDBMSs (e.g., PostgreSQL/PostGIS) in this era of big spatial data, and then reviewing the spatial support of popular NoSQL databases; (2) Data Processing: big data infrastructures, where big data processing systems are classified based on underlying architecture (e.g., Hadoop, Spark, NoSQL, and others) and the type of data processing system (such as spatial, spatio-temporal, trajectory, and spatial stream); (3) Data Programming and Software Tools, which summarizes available libraries, packages, and tools for processing spatial, spatio-temporal, and trajectory data in widely used programming languages, such as R and Python. Two popular GIS software, ArcGIS and QGIS, are also discussed in this last category.

The rest of the paper is organized as follows. Section~\ref{sec:related_works} provides an overview of existing surveys along with limitations, thus setting the goals of this survey. Section~\ref{sec:st_applications} discusses the importance of spatio-temporal data analytics research, along with a few important application domains. Spatial and spatio-temporal data is defined in Section~\ref{sec:st_what}. Section~\ref{sec:st_sql_nosql} defines the significance of spatial RDBMSs in the era of big spatial data and discusses the spatial support of NoSQL systems. The detailed review of existing big data infrastructures for processing spatial and spatio-temporal data is presented in Section~\ref{sec:st_big_data_sys}. Section~\ref{sec:st_prog_env} presents the ecosystem containing libraries, packages, and tools of two popular languages, Python, and R for processing spatial, spatio-temporal, and trajectory data along with two popular GIS software, ArcGIS and QGIS. Finally, Section~\ref{sec:conclusion} concludes the paper.             
\section{Related Works}
\label{sec:related_works}

A number of research works have been published which surveyed existing spatial data analytic systems. These surveys either performed a comparative analysis of existing spatial data analytics based on supported features or evaluated the performance of the existing systems by running supported spatial queries. Therefore, we divided the existing works into two groups, (1) surveys, (2) performance analyses. 

\textbf{Existing surveys:} In their survey of big spatial data systems, Eldawy et al.~\cite{Eldawy2015_BigSpatial_Survey} have explored the existing works that are developed on or before 2016, based on six key features. These features include (i) the system implementation approach, which defines whether the system was built as a library on-top of a system, built inside the core of a system, or developed as a new system from scratch, (ii) underlying data processing architecture (such as MapReduce, key-value stores, parallel DBMS, spark and other architectures), (iii) supported query language, (iv) supported indexing techniques, (v) supported spatial operations, and (vi) data visualization. This survey also provides an overview of some applications of spatial data. Maguerra et al.~\cite{Maguerra2018_ST_Survey} survey  provides a comprehensive review of big spatio-temporal data processing systems in the context of the underlying processing frameworks, partitioning and indexing techniques, and supported spatial queries. Castro et al.~\cite{Castro2018_DistSpatial_Survey} survey  analyzes the supported features of Hadoop and Spark-based spatial data systems from the user’s viewpoint to help users to select spatial data processing systems for their applications. Yao et al.~\cite{Yao2018_Spatial_VSurvey} studied and discussed recent technologies and techniques for big spatial vector data management based on the data model, storage, indexing, and processing and analysis. Karim et al.~\cite{Karim2018_ST_Aspects} portrait the spatio-temporal aspect of big data and performed a comparison of the supported spatio-temporal features on different frameworks, such as Apache Hadoop~\cite{url:hadoop} (SpatialHadoop), Apache Samza~\cite{Noghabi:2017:Samza}, Apache Storm~\cite{url:Apache-Storm}, Apache Spark~\cite{url:spark} (SpatialSpark and GeoSpark), and Apache Flink~\cite{Friedman:2016:Apache_Flink}. Almeida et al.~\cite{deAlmeida2020_SurveyTrajAnalytics} presented a survey on big trajectory data analytics from the viewpoints of storage, processing, summarization, and analysis of trajectories. This survey also provides an overview of a few systems for processing big trajectory data along with traditional systems based on PostgreSQL/PostGIS, Oracle Spatial, and other databases. The big trajectory systems include a cloud-based system on Microsoft Azure~\cite{Bao2016_TrajSystems}, ST-Hadoop~\cite{Alarabi2018_ST-Hadoop}, TrajSpark~\cite{Zhang2017_TrajSpark}, DiStRDF~\cite{Nikitopoulos2018_DiStRDF}, and systems based on Apache Flink, MongoDB~\cite{url:mongodb}, and other databases for semantic trajectories. Recently, Guo et al.~\cite{Guo2020_SpatialNoSQL_Survey} have surveyed the geospatial data processing capabilities in the 10 most popular NoSQL databases based on supported geometry types, geometry functions, spatial indexes, query languages, and data formats. This survey also discussed the strengths and weaknesses of each of these databases. 

\textbf{Existing performance analyses:} Hulbert et al.~\cite{Hulbert2016_BigSpatial_PStudy} performed an experimental study on GeoMesa~\cite{url:geomesa} and Elasticsearch~\cite{url:elasticsearch} by running spatio-temporal queries, where the authors have compared these systems based on query execution time and throughput. García et al.~\cite{Francisco2017_BigSpatial_Comparison} conducted a comparative analysis of the performance of SpatialHadoop~\cite{Eldawy2015_SpatialHadoop} and LocationSpark~\cite{Tang2016_LocationSpark} based on parallel and distributed spatial distance join queries. Hagedorn et al.~\cite{Hagedorn2017_BigSpatial_PStudy} performed feature comparison and performance analysis of Hadoop and Spark-based big spatial data processing systems. They have conducted a performance evaluation by running range and spatial join queries on SpatialHadoop, SpatialSpark~\cite{You2015_SpatialSpark}, GeoSpark~\cite{Yu2015_GeoSpark1}, and STARK~\cite{Stefan2017_STARK}, where STARK is proposed by the authors as a spatio-temporal extension of Spark. Data Reply~\cite{Reply2017_SpatialBenchmark} published a report on benchmarking six big geospatial data infrastructures to help the users to select the right infrastructures for their applications. These infrastructures include GeoSpark, Hive~\cite{Ashish2009_Hive}, MongoDB, GeoMesa, Elasticsearch, and Postgres-XL~\cite{url:postgres-xl}. These systems are evaluated by running several queries (range, regular expression, and join queries) on three different datasets, where each dataset contains 10 billion records. This report  also provided performance tuning tips for each of the evaluated systems. Pandey et al.~\cite{Pandey2018_BigSpatial_PStudy} performed a comprehensive study to analyze available features of selected Hadoop and Spark-based spatial data processing systems. They also evaluated the performance of five Spark-based spatial data systems (SpatialSpark, GeoSpark, Magellan~\cite{url:magellan}, LocationSpark, and Simba~\cite{Xie2016_Simba}) based on supported spatial operations, which include range and kNN query, spatial join, distance join, and kNN join. The cost of memory is also considered for performance evaluation. Alam et al.~\cite{Mahbub2018_BigSpatial_PStudy} also performed a comprehensive feature analysis and performance evaluation on Hadoop and Spark-based spatial data processing systems. However, instead of evaluating limited supported spatial features, they have implemented OGC-compliant join predicates and analysis features on SpatialHadoop and GeoSpark to assess the performance by running a number of spatial join queries, spatial analysis queries, and range queries on a cluster of nodes using real-world datasets. The authors also included $SpatialIgnite$ as part of the evaluation, which is developed as extended spatial support for another in-memory computing system, Apache Ignite~\cite{url:ignite}. More et al.~\cite{More2018_BigSpatial_PStudy} performed an experimental study on SpatialHadoop and GeoSpark to evaluate the performance based on various geospatial tasks such as data compression, indexing, and kNN and range query on a single node computer, which is not an ideal scenario for evaluating big spatial data systems. Recently, Haynes et al.~\cite{Haynes2020_BenchBigRaster} proposed a benchmark to evaluate spatial raster operations on big data platforms. They have assessed the performance of five raster operations on three big data platforms, namely PostgreSQL/PostGIS~\cite{url:PostGIS}, SciDB~\cite{Becla2013_SciDB}, and GeoTrellis~\cite{url:GeoTrellis}, using three different datasets. These operations include pixel count, reclassification, raster add, focal operations, and zonal statistics.

\textbf{Limitations of existing surveys and performance analyses:} First, existing surveys and performance analyses are not up-to-date. These researches mostly covered systems that were developed on or before 2017. Second, most of these works analyze the big data infrastructures for processing spatial data. However, in the meantime, a number of big data processing systems for spatio-temporal and trajectory data emerged in addition to a few new big spatial data processing systems. Though Guo et al.~\cite{Guo2020_SpatialNoSQL_Survey} have reviewed native spatial supports of NoSQL databases, there is no comprehensive survey of big spatial data processing systems, which are developed by utilizing NoSQL databases. Moreover, there is no review available on Python libraries such as DASK and RAPIDS for big spatial data processing. However, at present, along with Spark, these parallel and distributed libraries are gaining popularity and often considered as the next big data processing platform. Finally, researchers have not considered the other parts of the ecosystem of spatio-temporal analytics, namely spatial RDBMSs, GIS software, and spatial support in programming languages for processing spatial, spatio-temporal, and trajectory data. 

\textbf{Goal of this survey:} The goal of this survey is to conduct a comprehensive review on the current state of spatio-temporal data analytics systems research for processing spatial, spatio-temporal, and trajectory data. This survey discusses the up-to-date spatial support in RDBMSs, NoSQL databases, big data processing platforms, programming languages, and GIS software. First, along with the review of spatial support in popular relational (SQL) and NoSQL databases, we have addressed the significance of relational spatial databases in this era of big spatial data. Second, since a large number of research works have contributed by incorporating spatial support in big data platforms, we have studied and discussed these platforms based on the type of supports, such as spatial (both vector and raster), spatio-temporal, trajectory, or spatial streams. This survey also includes two new emerging big data processing platforms, DASK and RAPIDS, in addition to Hadoop, Spark, and NoSQL databases. Till now, there is no survey on spatial support in programming languages. This survey provides an overview of available libraries, packages, and tools of R and Python for processing spatial, spatio-temporal, and trajectory data. The APIs for interfacing R and Python with spatial RDBMSs, GIS software, and big data platforms are also discussed. Besides, a summary of spatial supports in other popular programming languages (e.g., C/C++, Java) is provided. Finally, this survey presents a review of two popular GIS software, ArcGIS and QGIS. We hope that this survey will help researchers, developers, and other stakeholders towards furthering the state-of-the-art.     
\section{Importance and Applications of Spatio-temporal Data Analytics}
\label{sec:st_applications}

In 1854, nearly 10 percent of a neighborhood at 40 Broad Street of the city of London died in just seven days due to the severe outbreak of deadly disease Cholera. Dr. John Snow~\cite{url:JohnSnow}, a British Physician, was able to identify the source (a water pump at Broad Street) of the disease by plotting cases of Cholera on a map called ghost map~\cite{url:cholera_outbreak}. This deadly map has taught us that the inherent knowledge of a map can solve a problem. Now, if we look at the current coronavirus (COVID-19) situation, or last year's bushfire situation in Australia, or the aftermath of recent floods/hurricanes around the world, or the effects of climate change, we observe that a large volume of people were affected by these deadly events. Forests were brunt, and animals lost their habitats. Though we are more technologically capable than ever, we are facing these situations or problems with more strength and more frequently. Therefore, it is essential to utilize the knowledge of spatial and temporal properties of data to mitigate or tackle many of these problems. The outcome of mined information from spatial and spatio-temporal data has already touched many avenues of human endeavor. Due to  recent technological advancements, location-based services and applications are an integral part of our daily activities. Researchers have been using these data for urban planning, navigating vehicles, identifying road accidents, tracking the activity of diseases (e.g., flu) and natural phenomena (e.g., hurricanes, tornados), and solving many other problems. In this context, it is important to put a closer look into the problems we are facing today and look beyond the traditional application domains for tackling future adverse events. In this section we discuss some important application domains, which need special attention now and in the future.

The ongoing COVID-19 virus situation came as a shock and has stopped the pace of the world. People lost their jobs and businesses, healthcare systems are overwhelmed with patients, many countries are struggling to give minimum services to emergency patients, and governments are facing difficulties to maintain supplies of food and necessary medical equipment. Even technologically advanced countries like USA and UK are not able to tackle this pandemic efficiently. Spatio-temporal data analysis, visualization, and mapping in the domain of \textbf{epidemiology} and \textbf{public health} may become essential tools for tackling future pandemics. A number of research works have  discovered spatio-temporal patterns~\cite{Matsubara2014_Epidemiology} and the spread of the diseases by studying the patient's treatment history. However, along with technological support, there is a need to utilize the spatio-temporal tools and techniques for efficient studies of the pandemics to quickly find the originating point, to stop spreading and isolating patients, and to provide better health-care services. In addition, other diseases like HIV, influenza, malaria, dengue, zika, and many other viruses are also a constant threat to the health sector around the world.

Similarly, the 2019-2020 season bushfire in Australia was the biggest in Australian history~\cite{news:Sydney2019}. As of March 2020, the bushfire burnt more than 18 million hectors across the country, destroyed over 5900 buildings (including more than 2000 homes), killed at least 34 people, 3 billion animals have been killed, and some animals may be driven to extinction~\cite{news:mail2020, news:canberratimes2020}. According to NASA, 306 million tones of CO\textsubscript{2} were emitted during the 2019-2020 Australian bushfire season~\cite{news:Time2020}. Besides, countries like USA, Canada, and a few European countries are also affected by bushfires every year. Spatio-temporal analysis of \textbf{bushfires} based on aerial and satellite images is useful to spot and tackle the bushfires at the initial stage to save forests and the lives of people and animals. Also, scientists have warned that if it is not possible to control the emission of greenhouse gas, bushfires could become a normal scenario every year~\cite{news:Reuters2020}. On the other hand, due to cyclones, hurricanes, and floods, a huge volume of people were affected and died around the world every year. Even the world's most developed and technologically equipped country, the USA, was adversely affected by Tropical Storm Nestor and Hurricane Dorian in 2019. Thus, \textbf{climatology} is another pivotal field to discover spatio-temporal patterns and relationships of climate variables and helps to prepare to tackle future adverse conditions~\cite{2018Atluri_STMining}. Since pollution (air, water, sound, and other pollution's) is an ongoing issue for a long time, the study of the \textbf{environmental science} domain to discover factors of pollution using data collected from sensors will always be as crucial as today.   

A large number of people are affected by an increasing number of large scale crises and disasters every year, such as hurricanes, bushfires, floods, earthquakes, epidemics, and other emergency events. These also can be small scale local emergency events such as road accidents, crimes, and house fires. During any such events, immediate actions are required to mitigate the suffering of people. These actions include to rescue and alert peoples, to maintain the supply chains (foods, medical supplies, and other essential resources), to provide medical services, and other necessary steps. Currently, GIS tools and solutions~\cite{Scholten2008_Emergency, Bhanumurthy2015_Emergency, Rifaat2016_Emergency, Bhanumurthy2017_Emergency} are used by emergency teams around the world to analyze data collected from aerial drones, satellites, smartphones, social networks, and other sources to take immediate measures. Future geospatial analytics will be more advanced and accurate for \textbf{emergency management and response} due to the integration of AI and machine learning. 

As the \textbf{ocean and marine environment} is the largest part of the earth, spatio-temporal ocean and marine datasets collected from widespread sources (such as satellites, remote sensors, aerial drones, stations, ships, buoys, and underwater sensors) are valuable in many application domains. These domains include safe and secure maritime navigation, autonomous cargo shipping, aquaculture production optimization, improved detection and forecasting of environmental changes, advanced weather forecasting, anomaly detection for identifying smuggling or drug trafficking, maritime surveillance, classifying acoustic sounds, and more. In addition, maritime shipping is the backbone of world trade and manufacturing supply chains. Therefore, application domains related to the ocean and marine environment are hot fields of research~\cite{Wright_Ocean_GIS}.

Due to the advancement of GPS technology and the internet, most taxis are equipped with a GPS device in large cities, and people use online location-based services, such as Uber, Google Maps, Foursquare, and other services for traveling purposes. A taxi driver always wants to get a passenger quickly and maximize the
profit. Whereas, a passenger wants to reach destination as quickly as possible. A number of research works~\cite{Yuan2011_Passenger, yuan2012t-finder, Meng2014_R-Taxi, Garg2018_R-Taxi} have been done in the last couple of years intending to maximize the profit of taxi drivers and utilize the valuable time of passengers. The outcome of these researches was already implemented into location-based services we are using today. These services can recommend locations for taxi drivers to find passengers quickly and passengers to find a taxi on time, which in turn reduce energy consumption and air pollution as well. This research also helps traffic engineers to implement policies to reduce traffic congestion and identify road accidents. However, as we are moving towards driver-less taxi services,  \textbf{intelligent transportation systems} is becoming an important field of research. On the other hand, researchers also started utilizing data generated from urban areas (such as sensors, vehicles, and humans) for \textbf{urban design and planning}~\cite{Yuan2012_Urban, Yuan2015_Urban}, which is another domain to watch-out.

We need to increase agricultural productions to ensure food security for the growing population of the world. However, the arable land area is decreasing due to the high growth of population and urbanization. Also, the fertility of lands is reducing as a result of the excessive use of fertilizers, pesticides, herbicides, water, and other inputs. Besides, crop production is affected by floods, drought, soil erosion, and other calamities. On top of that, farmers might lose farmlands due to the rise of sea levels as an effect of climate change. Increasing crop production as well as reducing the cost of production and unnecessary use of fertilizers, pesticides, and other farm inputs are very challenging tasks. \textbf{Precision agriculture}~\cite{Grisso2005_PreAgriculture, Esri2008_PreAgriculture, Delgado2019_PreAgriculture} uses knowledge extracted from spatio-temporal data collected from aerial drones, satellites, sensors, and other sources to identify soil types, crop diseases, and other attributes. This information helps farmers to identify site-specific needs and optimal use of farm inputs such as fertilizers, pesticides, water, and so on to maximize crop production and profits. Therefore, precision agriculture is another significant field of research that requires more attention from the spatio-temporal research community.

Other application domains, such as \textbf{Animal Migration and Forestation}, \textbf{Space Exploration}, and \textbf{Neuroscience}, are also prominent fields of research. Researchers also suggested exploring beyond the traditional application domains, such as \textbf{biology}, \textbf{chemistry}, \textbf{astronomy}, and more~\cite{Wang2020_SIComm}.    
\section{Definition and Types of Spatial and Spatio-temporal Data}
\label{sec:st_what}
Currently available spatial databases, big spatial data processing infrastructures, programming languages, and software tools have built support to model, store, and process either spatial data or spatio-temporal data. Spatio-temporal data can be either discrete point data or trajectory data. This section will define and differentiate among these types of data. 

A data item related to space (location-aware or geo-tagged) is called $geospatial$ or $spatial$ data. Traditionally, raster data (e.g., satellite images), point data (e.g., crime reports), or network data (e.g., road maps) were known patterns of spatial data~\cite{Evans2018}. In recent years, the traditional pattern of spatial data has changed due to the wide adoption of GPS-enabled mobile devices and the popularity of location-based services (LBS) and applications. Examples of this change include check-ins, GPS trajectories of smartphones, geo-tagged tweets, Instagram or Flickr photos, and so on. Spatial data types can be divided into three categories: vector, raster, and network data.

\textbf{Raster} data is represented as a collection of pixels (or grid cells), where each pixel is associated with a specific geographical location. Raster data can be discrete (such as land-cover type, soil type) or continuous (such as temperature, elevation, aerial photographs, satellite images). \textbf{Vector} data can be represented by points (e.g., a city, a movie theater), lines (e.g., roads, rivers, cables for phone or electricity) or polygons (e.g., a country, a lake, a river, a national park). A \textbf{spatial network} is a special graph that consists of nodes embedded in space. The most common example of a spatial network is the transportation network (e.g. the road network), where edges represent road segments, and nodes represent the intersection of road segments or points of interest~\cite{Huang2009_Spatial-Network}.

On the other hand, spatial data is being captured with a timestamp (temporal-tag) called spatio-temporal data, i.e., spatio-temporal data contains both spatial and temporal aspects of an object. Spatio-temporal data can also be defined as geometries changing over time~\cite{Erwig1999_ST-Types}. There are a number of data models (such as event model, temporal snapshot model, temporal change model and more) to represent spatio-temporal data in data processing systems~\cite{Pelekis2004_ST-Models, 2015Shekhar_STMining}. Spatio-temporal data type is basically the integration of timestamps (e.g., time instance, period, interval) with the spatial data type (e.g., point, line, polygon). Several classes of spatio-temporal data types are available in real-life application domains to represent an object with respect to both space and time. Kisilevich et al.\cite{Kisilevich2010_ST-Clustering} have considered point objects and defined five classes of spatio-temporal types, which include spatio-temporal events, geo-referenced variables, geo-referenced time series, moving objects, and trajectories. Whereas, Atluri et al.~\cite{2018Atluri_STMining} have described four classes of spatio-temporal data types, such as event data, point reference data, trajectory data, and raster data.


When raster data is collected with a timestamp, it is called \textbf{spatio-temporal raster} data. For example, air quality observations data from ground-based sensors or earth surface observations data from satellites are raster spatio-temporal data~\cite{2018Atluri_STMining}. In both of these cases, data is collected at fixed locations in space over time. The classification of vector data with timestamps depends on the type of geometries, such as points, lines, and polygons (regions)~\cite{Erwig1999_ST-Types}.

If we consider spatial points and point of time, spatio-temporal data can be either discrete point data or trajectory data. \textbf{Spatio-temporal discrete point} data can be event data that represent where and when the event happened. For example, a traffic accident can be represented by accident location and time of the accident. Therefore, event data can be used to model many real-life events such as crime events, disease outbreaks, road accidents, plane crashes, volcano eruptions, and more~\cite{2018Atluri_STMining}. Spatio-temporal discrete point data can also be point reference data, which is collected from a set of moving reference points in space over time. For example, drone observations of bushfires at point locations in space over time is discrete point reference data~\cite{2018Atluri_STMining}. Whereas, a \textbf{trajectory} is a path that consists of a set of points generated by moving objects in geographical space over time. The main sources of trajectory data are either GPS-enabled devices (e.g., taxi trajectories) or sensors attached to moving objects (e.g., animal trajectories). Trajectories can be classified into four main categories, such as mobility of humans, vehicles, animals, and natural phenomena (such as hurricanes, tornados)~\cite{Zheng2015_TrajReview}. The knowledge derived from trajectory data is important in many application domains, such as intelligent transportation systems, urban planning, location-based social networks, recommendation systems, animal migration analysis, and more.

Similarly, spatio-temporal data can be classified by considering other geometries like lines and regions instead of points, and timestamps like interval and period instead of a point of time~\cite{Erwig1999_ST-Types, Kisilevich2010_ST-Clustering}. Finally, due to the nature of spatio-temporal data, we can perform queries based on spatial, temporal, and spatio-temporal properties and relationships.
\section{Spatial Databases}
\label{sec:st_sql_nosql}

Spatial databases can be  divided into two main categories, relational databases (SQL), and NoSQL databases. Traditional relational database management systems (RDBMSs) have been around and serving us well for a long time. These RDBMSs with spatial support are stable, mature, efficient and have been used in a wide range of application domains at the enterprise level. Due to the large volume and diverse form of data being generated from a wide range of sources, NoSQL database systems, and big spatial data processing platforms have emerged. One may ask what is the significance of Spatial RDBMSs in this era of big spatial data. But Spatial RDBMSs are adapting to this era by integrating new features continuously. Researchers from industry and academia developed a few parallel and distributed systems by utilizing spatial RDBMSs. In this context, researchers are still using these modern spatial RDBMSs in a wide range of application domains. This section will address the significance of modern spatial RDBMSs in this era of big spatial data and explores the spatial support of both SQL and NoSQL databases.


\subsection{Spatial Relational Databases}
\label{sub-sec:st_RDBMS}

Traditional RDBMSs are popular for efficient data management and query processing. Therefore, research and development of spatial and spatio-temporal database systems have started by adding  extensions to traditional RDBMSs. For example, PostgreSQL/PostGIS~\cite{url:PostGIS}, Oracle Spatial~\cite{url:OracleSpatial}, IBM DB2 Spatial Extender~\cite{David2001_DB2Spatial}, Microsoft SQL Server~\cite{Fang2008_SQL_Server_Spatial}, MySQL Spatial~\cite{MySQL_Manual_2008}, and SpatialLite~\cite{url:SpatiaLite} are some popular spatial RDBMSs. The up-to-date features of these popular spatial RDBMSs are summarized in Table~\ref{tab:Spatial_RDBMSs}. These spatial RDBMSs are mature, stable, and contain  efficient SQL query engines. All of these systems support data formats (WKT and WKB) and geometry objects (point, linestring, polygon, and collections) specified by OGC Simple Features for SQL (part-2)~\cite{url:OGC}. Also, most of these spatial RDBMSs support R-Tree type indexing, except SQL Server and IBM DB2, where grid indexing has been utilized. Among them, only PostgreSQL/PostGIS and Oracle Spatial can store and process spatial raster data. Popular spatial RDBMSs such as PostgreSQL/PostGIS, Oracle Spatial, and SQL Server provide the complete set of spatial relationship and analysis functions defined in OGC~\cite{url:OGC} and ISO SQL/MM (part-3)~\cite{Stolze2003_SQL/MM} standard. Therefore, a wide range of spatial queries (e.g., spatial join, range) can be executed in these databases. 

\begin{table}[ht!]
\centering
\caption{Popular Spatial RDBMSs}
\label{tab:Spatial_RDBMSs}
\begin{tabular}{|c|c|c|c|c|c|}
\hline
  &
  \textbf{\begin{tabular}[c]{@{}c@{}}Data\\Formats\end{tabular}} &
  \textbf{\begin{tabular}[c]{@{}c@{}}Geometry\\Types\end{tabular}} &
  \textbf{\begin{tabular}[c]{@{}c@{}}Spatial\\Indexing\end{tabular}} &
  \textbf{\begin{tabular}[c]{@{}c@{}}Raster\\Support\end{tabular}} &
  \textbf{\begin{tabular}[c]{@{}c@{}}Spatial\\Functions\end{tabular}} \\ \hline
  
\textbf{\begin{tabular}[c]{@{}c@{}}PostgreSQL/\\PostGIS\end{tabular}} &
  \begin{tabular}[c]{@{}c@{}}WKT, WKB,\\ GML, KML,\\ GeoJSON,\\ SVG\end{tabular} &
  \begin{tabular}[c]{@{}c@{}}Point,\\ LineString,\\ Polygon,\\ Collections\textsuperscript{1}\end{tabular} &
  \begin{tabular}[c]{@{}c@{}}GiST,\\ SP-GiST,\\ BRIN\end{tabular} &
  Yes &
  \begin{tabular}[c]{@{}c@{}}OGC SFA-SQL,\\ ISO SQL/MM\textsuperscript{2}\end{tabular} \\ \hline
  
\textbf{Oracle Spatial} &
  \begin{tabular}[c]{@{}c@{}}WKT, WKB\\ JSON,\\ GeoJSON\end{tabular} &
  \begin{tabular}[c]{@{}c@{}}Point,\\ LineString,\\ Polygon,\\ Collections\end{tabular} &
  R-Tree &
  Yes &
  \begin{tabular}[c]{@{}c@{}}OGC SFA-SQL,\\ ISO SQL/MM\end{tabular} \\ \hline
  
\textbf{\begin{tabular}[c]{@{}c@{}}Microsoft\\SQL Server\end{tabular}} &
  \begin{tabular}[c]{@{}c@{}}WKT, WKB,\\ GML, GeoJSON\end{tabular} &
  \begin{tabular}[c]{@{}c@{}}Point,\\ LineString,\\ Polygon,\\ Collections\end{tabular} &
  \begin{tabular}[c]{@{}c@{}}Multi-level\\Grid\end{tabular} &
  No &
  \begin{tabular}[c]{@{}c@{}}OGC SFA-SQL,\\ ISO SQL/MM\end{tabular} \\ \hline
  
\textbf{\begin{tabular}[c]{@{}c@{}}IBM DB2\\Spatial Extender\end{tabular}} &
  \begin{tabular}[c]{@{}c@{}}WKT, WKB, \\ GML, KML\\ ESRI Shapefile\end{tabular} &
  \begin{tabular}[c]{@{}c@{}}Point,\\ LineString,\\ Polygon,\\ Collections\end{tabular} &
  Spatial Grid &
  No &
  \begin{tabular}[c]{@{}c@{}}OGC SFA-SQL,\\ ISO SQL/MM\end{tabular} \\ \hline
  
\textbf{MySQL Spatial} &
  WKT, WKB &
  \begin{tabular}[c]{@{}c@{}}Point,\\ LineString,\\ Polygon,\\ Collections\end{tabular} &
  R-Tree &
  No &
  OGC SFA-SQL \\ \hline
  
\textbf{\begin{tabular}[c]{@{}c@{}}SQLite/\\SpatialLite\end{tabular}} &
  WKT, WKB &
  \begin{tabular}[c]{@{}c@{}}Point,\\ LineString,\\ Polygon,\\ Collections\end{tabular} &
  R*-Tree &
  No &
  OGC SFA-SQL \\ \hline
\end{tabular}
\raggedright \small \textsuperscript{1}Collections - MultiPoint, MultiLineString, MultiPolygon, GeometryCollection\\
\raggedright \textsuperscript{2}Support functions compliant with OGC SFA-SQL~\cite{url:OGC} and ISO SQL/MM~\cite{Stolze2003_SQL/MM} standard \\
\end{table}

However, due to the I/O bottleneck, lack of parallelism and scalability, the performance of these systems deteriorated with the increasing volume of data. Also, it is challenging to model heterogeneous and multidimensional data in spatial RDBMSs. Still these databases went through a lot of changes in the last couple of years. Researchers and developers are continuously integrating new features to these systems or utilizing these systems to meet the current demands of spatial data. For example, the GeoJSON data format of PostgreSQL/PostGIS. Therefore, this section will also address the significance of modern spatial RDBMSs in this era of big spatial data by discussing changes made in one of the most popular spatial RDBMS, PostgreSQL/PostGIS.

PostgreSQL is an open-source, vertically scalable, and extensible RDBMS. PostGIS is a spatial extension of PostgreSQL, which supports OGC-compliant spatial SQL queries. Vertical scaling can improve the performance of PostgreSQL/PostGIS on a single computer system, but horizontal scalability is required for processing a large volume of spatial data. We can achieve horizontal scalability through sharding in PostgreSQL. One can also achieve read scalability by utilizing pgpool (Pgpool-II~\cite{url:pgpool}) and streaming replication instead of sharding. However, sharding can reduce the I/O bottleneck significantly by partitioning data across multiple nodes of a cluster~\cite{url:postgres-sharding}. Several solutions are available where horizontal scalability and query parallelism achieved through sharding, such as Postgres-XL~\cite{url:postgres-xl}, Citus~\cite{url:citus}, PL/Proxy~\cite{url:plproxy}, etc. PostGIS can be integrated with both Citus and Postgres-XL. PostgreSQL (v9.6+) also has a built-in sharding feature called Foreign Data Wrappers (FDW), which allows PostgreSQL to access data from external sources. Therefore, data can be distributed across nodes of a cluster, where each partition can be accessible through FDW directly from disk or main memory instead of local tables of PostgreSQL. Moreover, features like parallel sequence scans, parallel joins, and parallel aggregates for parallel spatial query processing are now completely working with PostgreSQL (v12) and PostGIS (v3.0). Thus, one can take advantage of default parallel processing support to process large scale spatial data in PostgreSQL/PostGIS~\cite{url:parallel-postgis}.

As today's big data comes from diverse sources in different formats, it is not always possible to store data in a tabular format in RDBMS. Therefore, NoSQL database systems emerged in the last decade. However, JSON and JSONB data types were added to PostgreSQL in 2012 and 2014, respectively. Recently, SQL/JSON was introduced in PostgreSQL v12, which is compliant with the SQL-2016 standard. The SQL-2016 standard has recognized  NoSQL and includes features for the SQL/JSON data model and path language as well as commands for storing, publishing, and querying JSON data. Thus, now we can query and index JSON (JSONB) data in PostgreSQL~\cite{url:postgres-nosql}.

Some research work has also utilized PostgreSQL/PostGIS to develop parallel query processing infrastructures for spatial data~\cite{Alam2018_MCSThesis_BigSpatial}. For example, Niharika~\cite{Ray2011_Niharika} utilizes the powerful features of PostgreSQL/PostGIS to implement a parallel query processing system along with efficient data declustering and load balancing techniques. However, as its storage layer is not distributed, it needs to replicate the whole dataset in each node of a cluster. On the contrary, each node of Paragon~\cite{Haynes2015_Paragon} needs to host a subset of the partitions only. MobilityDB~\cite{Esteban2019_MobilityDB_1} was developed as an extension of PostgreSQL/PostGIS, which provides support for storing and querying moving objects data (spatial trajectory). This support includes spatio-temporal data types, indexing techniques, and query operations. Recently, MobilityDB emerged as a distributed system by integrating with Citus~\cite{url:citus} for processing massive trajectory data~\cite{Bakli2019_MobilityDB_2}. CARTO~\cite{url:carto} provides spatial analysis and mapping services for a wide range of application areas for many organizations by developing APIs, libraries, and tools. Developers and GIS scientists use CARTO for developing GIS applications. CARTO also uses PostgreSQL/PostGIS underneath as a spatial database server.

Similarly, other spatial RDBMSs (e.g., Oracle Spatial and SQL Server) are also incorporating new features continuously to adapt to this era of big spatial data. Besides, a large number of companies are still using spatial RDBMSs for their businesses. This means that currently, if someone ask the following questions: (i) can spatial RDBMSs (like PostgreSQL/PostGIS) scale well for the problems we are dealing today?; (ii) can these systems process massive datasets?; (iii) can these systems process data in different formats?; or (iv) how long will these systems survive in this era of big data?; the answers to these questions would be yes, these systems are scalable enough for many problems we are dealing with today and can process a certain volume of data in different formats, and will be around for a long time.


\subsection{Spatial NoSQL Databases}
\label{sub-sec:st_NoSQL}

NoSQL (Not-Only-SQL) database systems~\cite{Felix2017_NoSQL_Survey, Ali2018_NoSQL_Survey} are also known as non-relational database systems. Carlo Strozzi came with the NoSQL term in  1998~\cite{url:strozzi_NoSQL}. Since a large volume of data comes from diverse sources with various formats (such as semi-structured and unstructured), it is challenging to model these data using relational tables as there is no predefined fixed schema. Besides, traditional relational database systems suffer due to the lack of parallelism, I/O bottleneck, and horizontal scalability. Therefore, NoSQL database systems have emerged as an alternative data management technology in the last decade. NoSQL database systems can be classified into four broad groups based on the core data model, (1) Key-Value Databases (e.g., Redis, Oracle NoSQL), (2) Column Family (Wide-Column) Databases (e.g., Cassandra, HBase), (3) Document Databases (e.g., MongoDB, Couchbase), and (4) Graph Databases (e.g., Neo4j, ArangoDB).

NoSQL database systems are fault-tolerant, scalable, highly available, and support high update rates. However, a few of these systems have native spatial support currently. Spatial extensions have  been added to some of these NoSQL databases in recent years. The current spatial support of some popular NoSQL databases is presented in Table 2. Redis~\cite{url:redis} is an in-memory key-value store that has implemented a geohash spatial index to accelerate query processing. It operates based on a Geo Set data structure, which is built with a Sorted Set. A set of commands (e.g., geoadd, geopos, geohash, georadius, and more) are available in Redis to create an index and perform spatial operations on point datasets stored in Geo Sets. However, these spatial commands can only perform limited spatial analyses on point type geometry. Also, Redis does not support SQL-like query language. On the contrary, one can run SQL-like spatial queries on Oracle NoSQL~\cite{Oracle_NoSQL_2019}, which supports all common geometry objects, geohash indexing, and a set of spatial operators for processing spatial data.

\begin{table}[ht!]
\centering
\caption{Popular NoSQL Databases with Spatial Support}
\label{tab:NoSQL-Spatial}
\resizebox{\textwidth}{!}{%
\begin{tabular}{|c|c|c|c|c|c|c|c|}
\hline
\textbf{\begin{tabular}[c]{@{}c@{}}NoSQL\\ Database\end{tabular}} &
  \textbf{\begin{tabular}[c]{@{}c@{}}Data \\ Model\end{tabular}} &
  \textbf{\begin{tabular}[c]{@{}c@{}}Spatial\\ Support\end{tabular}} &
  \textbf{\begin{tabular}[c]{@{}c@{}}Data\\ Formats\end{tabular}} &
  \textbf{\begin{tabular}[c]{@{}c@{}}Geometry\\ Types\end{tabular}} &
  \textbf{\begin{tabular}[c]{@{}c@{}}Spatial\\ Indexing\end{tabular}} &
  \textbf{\begin{tabular}[c]{@{}c@{}}Spatial\\ Functions\end{tabular}} &
  \textbf{\begin{tabular}[c]{@{}c@{}}SQL-like\\Query Language\end{tabular}} \\ \hline
  
\textbf{Redis} &
  Key-Value &
  Native~\cite{url:redis-spatial} &
  GeoJSON &
  Point &
  geohash &
  \begin{tabular}[c]{@{}c@{}}geoadd, geodist,\\zrange, zscan,\\geopos, geohash,\\georadius,\\georadiusbymember,\\zrem\end{tabular} &
  \begin{tabular}[c]{@{}c@{}}Not\\Supported\end{tabular} \\ \hline
  
\textbf{\begin{tabular}[c]{@{}c@{}}Oracle \\ NoSQL\end{tabular}} &
  Key-Value &
  Native~\cite{Oracle_NoSQL_2019} &
  GeoJSON &
  \begin{tabular}[c]{@{}c@{}}Point,\\ LineString,\\ Polygon,\\ Collections\textsuperscript{1}\end{tabular} &
  geohash &
  \begin{tabular}[c]{@{}c@{}}geo\_intersect,\\ geo\_inside,\\ geo\_within\_distance,\\ geo\_near\end{tabular} &
  \begin{tabular}[c]{@{}c@{}}Yes\\ Supported\end{tabular} \\ \hline
  
\textbf{MongoDB} &
  Document &
  Native~\cite{url:mongodb} &
  \begin{tabular}[c]{@{}c@{}}GeoJSON,\\Legacy\\Coordinate\\Pairs\end{tabular} &
  \begin{tabular}[c]{@{}c@{}}Point,\\ LineString,\\Polygon,\\Collections\end{tabular} &
  \begin{tabular}[c]{@{}c@{}}2dsphere,\\ 2d\end{tabular} &
  \begin{tabular}[c]{@{}c@{}}\$near, \\ \$nearSphere,\\ \$geoWithin, \\ \$geoIntersect,\\ \$geoNear\end{tabular} &
  \begin{tabular}[c]{@{}c@{}}Not\\ Supported\end{tabular} \\ \hline
  
\textbf{Couchbase} &
  Document &
  GeoCouch~\cite{url:geocouch} &
  GeoJSON &
  \begin{tabular}[c]{@{}c@{}}Point,\\LineString,\\Polygon,\\Collections\end{tabular} &
  R-Tree &
  BBox &
  N1QL \\ \hline
  
\textbf{Cassandra} &
  Wide\_Column &
  \begin{tabular}[c]{@{}c@{}}Lucene Index\\Plugin~\cite{url:lucene-index}\end{tabular} &
  WKT &
  \begin{tabular}[c]{@{}c@{}}Point,\\ LineString,\\Polygon,\\Collections\end{tabular} &
  \begin{tabular}[c]{@{}c@{}}Lucene\\Index\\(Secondary\\Index)\end{tabular} &
  \begin{tabular}[c]{@{}c@{}}intersects,\\contains,\\is\_within\end{tabular} &
  CQL \\ \hline
  
 \multirow{2}{*}{\textbf{Neo4j}} &
      \multirow{2}{*}{Graph} &
      Native &
      N/A &
      \multicolumn{1}{l|}{Point (2D, 3D)} &
      Hilbert-curve &
      Distance &
      \multirow{2}{*}{Cypher} \\ \cline{3-7} &
    &
      \begin{tabular}[c]{@{}c@{}}Neo4j-Spatial\\~\cite{url:Neo4jSpatial}\end{tabular} &
      WKT, WKB &
      \begin{tabular}[c]{@{}c@{}}Point,\\ LineString,\\ Polygon,\\ Collections\end{tabular} &
      R-Tree &
      \begin{tabular}[c]{@{}c@{}}Contain, Cover, \\ Cross, Disjoint,\\ Intersect, Overlap,\\ Touch, Within, etc.\end{tabular} & \\ \hline
      
\end{tabular}%
}
\raggedright \tiny \textsuperscript{1}Collections - MultiPoint, MultiLineString, MultiPolygon, GeometryCollection\\
\end{table}

MongoDB~\cite{url:mongodb} is a document database that has native support for processing spatial data. MongoDB supports common GeoJSON objects (such as point, linestring, polygon, and collections) and \textit{2dsphere} indexes to model geometries on a spherical surface. It can also store geometries on a 2D surface as legacy coordinate pairs and \textit{2d} indexes to model 2D queries. MongoDB provides a set of operators such as \textit{\$near}, \textit{\$nearSphere}, \textit{\$geoWithin}, \textit{\$geoIntersect}, and \textit{\$geoNear} to perform spatial queries. However, like Redis, it does not have support for SQL-like queries. Whereas, Couchbase~\cite{url:couchbase} is another popular document database that supports SQL-like query language, N1QL. GeoCouch~\cite{url:geocouch} is a spatial extension for both Couchbase and Apache CouchDB~\cite{url:CouchDB}. GeoCouch is developed based on R-Trees and supports common GeoJOSN objects like MongoDB. It allows executing spatial queries using bounding-boxes (BBox). However, MongoDB is richer in terms of support to perform a wide range of spatial queries.

Apache Cassandra~\cite{url:cassandra} is a column family database that does not have native support for processing spatial data. Stratio's Lucene Index~\cite{url:lucene-index} is a spatial plugin for Cassandra, whose spatial index is an extension of Cassandra's secondary indexes. The Lucene plugin provides a set of spatial predicates (intersects, contains, and is\_within) and transformation functions (buffer, convexhull, union, and more), which enable Cassandra to store, index, and process common spatial objects such as point, linestring, polygon, and collections.  Brahim et al.~\cite{Brahim2016_Cassandra_Spatial} have also extended CQL (Cassandra Query Language) to add spatial support with Cassandra, which includes geohash indexing and spatial queries (within\_circle, within\_polygon, and within\_path).

Neo4j~\cite{url:neo4j} is one of the most popular graph database systems, which supports an efficient query language, Cypher. Neo4j Spatial~\cite{url:Neo4jSpatial} is a library that facilitates Neo4j to store, index, and process spatial data. This library contains modules to import spatial data (ESRI Shapefile and OSM), and R-Tree indexing can be applied during import or later to stored data. It also supports a wide range of spatial functions (contain, cover, intersect, and so on) to perform spatial operations on common geometric objects (point, linestring, polygon, and collections). Besides, it wraps popular geospatial libraries, JTS and GeoTools, and therefore, one can utilize the functionalities of these libraries in Neo4j. However, Neo4j spatial is an external library, and hence, it is not highly scalable. Moreover, this library suffers when applications require high concurrency and need to handle a large volume of data. Therefore, Neo4j (v3.4) introduced two native data types, spatial (Point) and temporal (Date, Time, DateTime, Duration, and other types). This point type supports both 2D and 3D points and can be specified by either a geographic or cartesian coordinate system. Neo4j uses Hilbert-curve for indexing points (2D or 3D) and only supports spatial distance function. Similarly, it also provides indexes and functions to process temporal data.

Researchers have also developed several big spatial data processing systems by utilizing the capability of NoSQL databases in recent years that will be discussed in Section~\ref{sub-sec:nosql-systems}. The performance of NoSQL databases is evaluated and discussed by several researchers for spatial workloads~\cite{Kim2019_Bench_NoSQL_Spatial, Kanwar2019_Bench_NoSQL_Spatial}. Some of these performance analyses also involve comparisons with relational spatial databases~\cite{Baralis2017_SQLvsNoSQL_Spatial, Andre2018_SQLvsNoSQL_Spatial, Agarwal2016_MongoDBvsPostGIS, Dominik2019_MongoDBvsPostGIS, Makris2019_MongoDBvsPostGIS}.

\subsection{Future Research Directions}
\label{sub-sec:st_gis_future_directions}
The codebase of spatial RDBSMs is mature, stable, efficient, and easily extensible. Also, these systems support efficient SQL queries along with current distributed and parallel capability. By considering the current state of spatial RDBMSs, it would be a great idea to incorporate distributed storage like HDFS or utilize the main memory like Spark in distributed systems developed based on spatial RDBMSs.

Currently, the spatial support of NoSQL databases lacks available spatial operations compared to spatial RDBMSs. Also, a few of these databases do not have support for SQL-like spatial queries. In addition, we need to work to add support to store and process spatial raster and trajectory data in NoSQL databases. At present, the graph database, Neo4j (v4.0), can scale horizontally through sharding, and therefore, it will be interesting to see the performance of distributed graph databases like Neo4j for processing spatial data.
\section{Big Spatio-temporal Data Processing Infrastructures}
\label{sec:st_big_data_sys}
With the rise of big spatial and spatio-temporal data and its application domains, there is demand for highly scalable and distributed data processing systems to store, manage, and process the massive volume of data. Therefore, researchers from both academia and industry are working towards achieving these demands. The big spatial (or spatio-temporal) data processing systems that have been developed in the last couple of years are mainly based on MapReduce framework Hadoop~\cite{url:hadoop}, NoSQL databases~\cite{Felix2017_NoSQL_Survey, Ali2018_NoSQL_Survey}, and Spark~\cite{url:spark}. Most of these systems have built spatial or spatio-temporal support either by adding a layer on top existing systems or by extending the core of the existing systems. Besides, a number of big spatial or spatio-temporal data processing systems have been developed either from scratch or by utilizing data processing platforms other-than Hadoop, NoSQL, and Spark. For example, recently, Python libraries such as DASK~\cite{Team2016_Dask} and RAPIDS~\cite{2018_RAPIDS} have started to gain popularity as a big data platform. 

\begin{figure}[!htbp]
	\centering
	\includegraphics[width=0.75\linewidth, keepaspectratio]{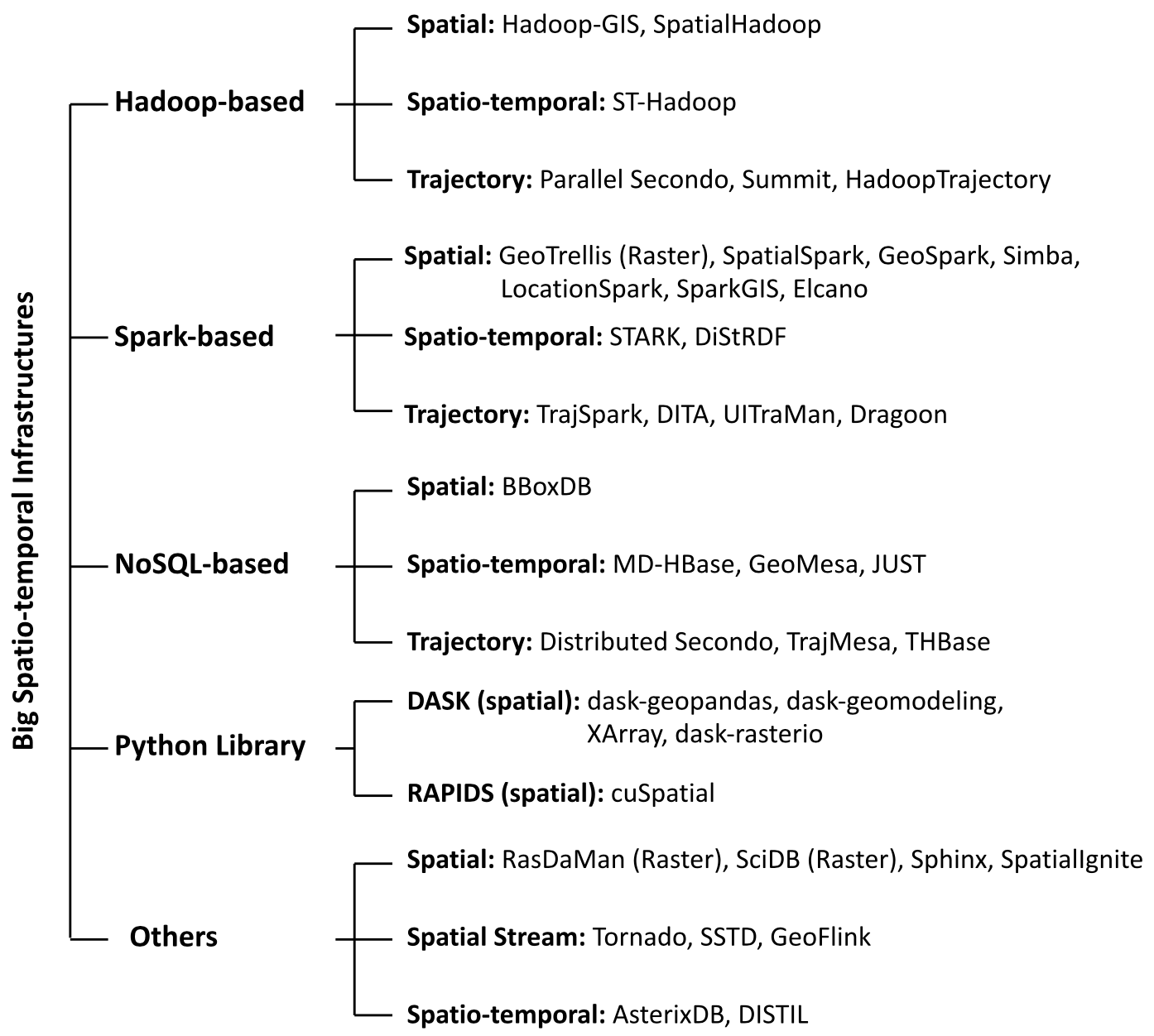}
	\caption{Overview of Big Spatio-temporal Data Processing Infrastructures}
	\label{fig:st_big_platforms}
\end{figure}

In this section, we first categorize the big spatio-temporal data infrastructures based on their development criteria. The systems under each of the categories are then divided into groups based on the type of data processing systems, which include spatial (spatial vector and raster, spatial stream), spatio-temporal, and trajectory. An overview of these systems is provided in Figure~\ref{fig:st_big_platforms}. Finally, this section provides a comprehensive review of each of these systems based on their supported features, such as data types, partitioning and indexing techniques, query language, and supported spatio-temporal operations.

\subsection{Hadoop-based Big Spatio-temporal Infrastructures}
\label{sub-sec:hadoop-systems}
Hadoop~\cite{url:hadoop} is a highly scalable and distributed open-source MapReduce~\cite{Dean2008_MapReduce} framework for processing a large volume of data, which is integrated with the HDFS~\cite{Shvachko2010_HDFS}  distributed storage system.  Hadoop does not have any native support for processing spatial (or spatio-temporal) data. Therefore, Hadoop distributes and indexes data across the clusters without considering the spatial (or spatio-temporal) aspect of data, which affects query processing performance on the data negatively. Due to the huge popularity of Hadoop as a big data processing framework in both research and industry communities, a number of extensions to Hadoop were proposed to store, process, and analyze spatial (or spatio-temporal) data. These systems include Hadoop-GIS~\cite{aji2013_hadoop-gis}, SpatialHadoop~\cite{Eldawy2015_SpatialHadoop}, ESRI Tools for Hadoop~\cite{Whitman2014_ESRIHadoop}, Parallel SECONDO~\cite{Ralf2015_PSecondo}, ST-Hadoop~\cite{Alarabi2018_ST-Hadoop}, Summit~\cite{Alarabi2019_Summit}, and HadoopTrajectory~\cite{Bakil2019_HadoopTrajectory}. A detailed feature matrix of these systems is provided in Table~\ref{tab:st-hadoop-systems}.

Hadoop-GIS~\cite{aji2013_hadoop-gis} is a spatial extension of Hadoop. It integrates a spatial layer on top of Hadoop instead of changing the core of the framework. As a result, the performance of Hadoop-GIS for processing spatial data is not quite as good as expected. Besides, Hadoop-GIS extends Hive~\cite{Ashish2009_Hive} to support declarative spatial querying (HiveSP) that adds an extra layer of overhead over Hadoop for processing spatial queries.  SpatialHadoop~\cite{Eldawy2015_SpatialHadoop} incorporates spatial support inside the core of the Hadoop framework. Therefore, it achieves better performance than Hadoop-GIS for running spatial queries on a large dataset. A SQL-like query language, Pigeon~\cite{Eldawy2014Pigeon}, which extends Pig Latin~\cite{olston2008_pig_latin}, is also introduced to run spatial queries on SpatialHadoop. However, the evaluation shows that Pigeon is not efficient to execute spatial join queries on large datasets~\cite{Mahbub2018_BigSpatial_PStudy}.

\begin{table}[ht!]
\centering
\caption{Hadoop-based Spatio-temporal Systems}
\label{tab:st-hadoop-systems}
\resizebox{\textwidth}{!}{%
\begin{tabular}{|c|c|c|c|c|c|c|}
\hline

  &
  \textbf{System Type} &
  \textbf{Data Types} &
  \textbf{Partitioning} &
  \textbf{Indexing} &
  \textbf{\begin{tabular}[c]{@{}c@{}}SQL-Like Query\\ Language\end{tabular}} &
  \textbf{\begin{tabular}[c]{@{}c@{}}Supported\\ Queries\end{tabular}} \\ \hline
  
\textbf{\begin{tabular}[c]{@{}c@{}}Hadoop-GIS~\cite{aji2013_hadoop-gis}\\ (2013)\end{tabular}} &
  Spatial &
  \begin{tabular}[c]{@{}c@{}}Point,\\ LineString,\\ Polygon\end{tabular} &
  \begin{tabular}[c]{@{}c@{}}SATO~\cite{Hoang2014_SATO}\\ Framework\textsuperscript{1}\end{tabular} &
  \begin{tabular}[c]{@{}c@{}}Two-Level\\(Global, Local)\\ R*-tree\end{tabular} &
  \begin{tabular}[c]{@{}c@{}}HiveSP\\ (Extended HiveQL)\end{tabular} &
  \begin{tabular}[c]{@{}c@{}}Range, kNN,\\ Join\end{tabular} \\ \hline
  
\textbf{\begin{tabular}[c]{@{}c@{}}SpatialHadoop~\cite{Eldawy2015_SpatialHadoop}\\ (2015)\end{tabular}} &
  Spatial &
  \begin{tabular}[c]{@{}c@{}}Point,\\ LineString,\\ Polygon\end{tabular} &
  \begin{tabular}[c]{@{}c@{}}Fixed-Grid,\\ STR\end{tabular} &
  \begin{tabular}[c]{@{}c@{}}Two-Level\\(Global, Local)\\ Grid File,\\ R-tree,\\ R+-tree\end{tabular} &
  \begin{tabular}[c]{@{}c@{}}Pigeon\\ (Extended Pig Latin)\end{tabular} &
  \begin{tabular}[c]{@{}c@{}}Range, kNN,\\ Join\end{tabular} \\ \hline
  
\textbf{\begin{tabular}[c]{@{}c@{}}Parallel SECONDO\\ \cite{Ralf2015_PSecondo}(2015)\end{tabular}} &
  \begin{tabular}[c]{@{}c@{}}Spatio-temporal\\ (Trajectory)\end{tabular} &
  \begin{tabular}[c]{@{}c@{}}Point, \\ LineString, \\ Region,\\ Instant, Period,\\ Periods, Interval\end{tabular} &
  3D Grid &
  \begin{tabular}[c]{@{}c@{}}B-Tree,\\ R-Tree\end{tabular} &
  Executable &
  Range, Join \\ \hline
  
\textbf{\begin{tabular}[c]{@{}c@{}}ST-Hadoop~\cite{Alarabi2018_ST-Hadoop}\\ (2017)\end{tabular}} &
  Spatio-temporal &
  \begin{tabular}[c]{@{}c@{}}STPoint,\\ Time,\\ Interval\end{tabular} &
  \begin{tabular}[c]{@{}c@{}}Time-Slice,\\ Data-Slice\end{tabular} &
  \begin{tabular}[c]{@{}c@{}}Two-Level\\ L1: Temporal\\ L2: Spatial\\ \\ Spatial: \\ SpatialHadoop\end{tabular} &
  Extended Pigeon &
  Range, Join \\ \hline
  
\textbf{\begin{tabular}[c]{@{}c@{}}Summit~\cite{Alarabi2019_Summit}\\ (2018)\end{tabular}} &
  \begin{tabular}[c]{@{}c@{}}Spatio-temporal\\ (Trajectory)\end{tabular} &
  \begin{tabular}[c]{@{}c@{}}Trajectory,\\ Time,\\ Interval\end{tabular} &
  \begin{tabular}[c]{@{}c@{}}Spatial-based,\\ Segmentation-based\end{tabular} &
  \begin{tabular}[c]{@{}c@{}}Two-Level\\ L1: Temporal\\ L2: Spatial\\ \\ Temporal: ST-Hadoop\\ Spatial: Extended\end{tabular} &
  Extended Pigeon &
  \begin{tabular}[c]{@{}c@{}}Range, kNN,\\ kNN Similarity,\\ Join\end{tabular} \\ \hline
  
\textbf{\begin{tabular}[c]{@{}c@{}}HadoopTrajectory\\ \cite{Bakil2019_HadoopTrajectory}(2019)\end{tabular}} &
  \begin{tabular}[c]{@{}c@{}}Spatio-temporal\\ (Trajectory)\end{tabular} &
  \begin{tabular}[c]{@{}c@{}}Point, Region,\\ Instant, Interval, \\ Periods,\\ TrajSegment, \\ Trajectory\end{tabular} &
  N/A &
  \begin{tabular}[c]{@{}c@{}}Grid, R-Tree\\ (3D Extension)\\ e.g., 3DR-Tree\end{tabular} &
  N/A &
  \begin{tabular}[c]{@{}c@{}}Range, kNN,\\ Join\end{tabular} \\ \hline
  
\end{tabular}%
}
\raggedright \tiny \textsuperscript{1}SATO supports Fixed-Grid, Binary-Space, Hilbert-Curve, Strip-based, and STR partitioning techniques.
\end{table}

Due to the lack of spatio-temporal data types, partitioning, and indexing techniques, both Hadoop-GIS and SpatialHadoop suffer when querying on spatio-temporal datasets. ST-Hadoop~\cite{Alarabi2018_ST-Hadoop} is a temporal extension of SpatialHadoop, which incorporates spatio-temporal awareness into each layer of SpatialHadoop. However, ST-Hadoop was developed by considering attributes of discrete spatio-temporal point data, not trajectory data, and properties of trajectory data are quite different from discrete point data. Therefore, if we partition and index  trajectory data using ST-Hadoop,  the performance of query processing will be be impacted negatively. For example, each individual trajectory of an object contains a set of points and an object can have multiple trajectories. Now, if we partition trajectories of a moving object using ST-Hadoop, they may be partitioned into different blocks of HDFS over different clusters, which will require more time to perform queries. Therefore, Summit~\cite{Alarabi2019_Summit} was developed as an extension of ST-Hadoop to include data types, partitioning and indexing techniques, and operations, for processing trajectory data. Bakli et al.~\cite{Bakil2019_HadoopTrajectory} have proposed HadoopTrajectory, which adds a diverse set of data types and operators into the core of Hadoop to store and process trajectory data. The careful integration of partitioning and indexing strategies for trajectory data into Hadoop layers makes their system an efficient big trajectory processing system. Parallel SECONDO~\cite{Ralf2015_PSecondo} integrates SECONDO DBMS with Hadoop for scalability. SECONDO~\cite{Guting2010_Secondo} is a prototype DBMS to store and process moving object data. SECONDO supports data models to represent spatial and temporal data, operations for processing moving objects, and a SQL-like query language. As SECONDO was developed for a single computer and unable to process big data, SECONDO was  combined with Hadoop to execute parallel queries and distribute data across a cluster of nodes. However, Hadoop only does the scheduling and query coordination tasks received from the SECONDO master node, while SECONDO executes the query in each worker node. Finally, SECONDO master node needs to aggregate the results. Therefore, Parallel SECONDO has not able to utilize the power of Hadoop properly due to its centralized behavior. 


\subsection{Spark-based Big Spatio-temporal Infrastructures}
\label{sub-sec:spark-systems}

Apache Hadoop is a disk-based system optimized for I/O efficiency. Therefore, the performance of Hadoop-based big data systems can deteriorate at scale. On the other hand, the growing main memory capacity in a cluster of machines has fueled the development of in-memory big data systems. Apache Spark~\cite{url:spark, Zaharia2010_Spark_1, Zaharia2016_Spark_2} is a popular and widely used distributed in-memory big data processing framework, which is implemented by taking advantage of a large pool of memory available in a cluster of machines to achieve better performance than disk-based systems. However, like Hadoop, Spark also does not have native support for processing spatial (or spatio-temporal) data. Due to the lack of spatial (or spatio-temporal) data types, partitioning and indexing strategies, and spatial operations, Spark process spatial (or spatio-temporal) data in the same way as non-spatial data. Therefore, several Spark-based spatial (or spatio-temporal) data processing systems have been developed in the last few years to alleviate these limitations. These systems include SpatialSpark~\cite{You2015_SpatialSpark}, GeoSpark~\cite{Yu2015_GeoSpark1}, LocationSpark~\cite{Tang2016_LocationSpark, Tang2019_LocationSpark}, Simba~\cite{Xie2016_Simba}, STARK~\cite{Stefan2017_STARK}, SparkGIS~\cite{Furqan2017_SparkGIS}, TrajSpark~\cite{Zhang2017_TrajSpark}, Elcano~\cite{Englinus2018_Elcano}, DiStRDF~\cite{Nikitopoulos2018_DiStRDF}, DITA~\cite{Zeyuan2018_DITA}, UlTraMan~\cite{Ding2018_UITraMan}, Dragoon~\cite{Fang2021_Dragoon}, and GeoTrellis~\cite{url:GeoTrellis}. The detailed feature matrix of these systems is presented in Table~\ref{tab:st-spark-systems}.

\begin{table}[hbt!]
\caption{Spark-based Spatio-temporal Systems}
\centering
\label{tab:st-spark-systems}
\resizebox{\textwidth}{!}{\begin{tabular}{|c|c|c|c|c|c|c|}
\hline
       &
      \textbf{System Type} &
      \textbf{\begin{tabular}[c]{@{}c@{}}Data\\Types\end{tabular}} &
      \textbf{Partitioning} &
      \textbf{Indexing} &
      \textbf{\begin{tabular}[c]{@{}c@{}}Query\\Language\end{tabular}} &
      \textbf{\begin{tabular}[c]{@{}c@{}}Supported\\Queries\end{tabular}} \\ \hline
      
\textbf{\begin{tabular}[c]{@{}c@{}}SpatialSpark~\cite{You2015_SpatialSpark}\\ (2015)\end{tabular}} &
      Spatial &
      \begin{tabular}[c]{@{}c@{}}Point,\\ LineString,\\ Polygon\end{tabular} &
      \begin{tabular}[c]{@{}c@{}}Fixed-Grid,\\ Binary-Split,\\ STR\end{tabular} &
      R-Tree &
      N/A &
      \begin{tabular}[c]{@{}c@{}}Range,\\ Broadcast-Join,\\ Partitioned-Join\end{tabular} \\ \hline
      
\textbf{\begin{tabular}[c]{@{}c@{}}GeoSpark~\cite{Yu2015_GeoSpark1}\\ (2015)\end{tabular}} &
      Spatial &
      \begin{tabular}[c]{@{}c@{}}Point,\\ LineString,\\ Polygon,\\ Rectangle\end{tabular} &
      \begin{tabular}[c]{@{}c@{}}Uniform-Grid\\ Voronoi,\\ R-Tree,\\ Quad-Tree,\\ KDB-Tree\end{tabular} &
      \begin{tabular}[c]{@{}c@{}}R-Tree,\\ Quad-Tree\end{tabular} &
      \begin{tabular}[c]{@{}c@{}}Extended\\ Spark SQL\end{tabular} &
      \begin{tabular}[c]{@{}c@{}}Range, kNN,\\ Spatial-Join,\\ Distance-Join\end{tabular} \\ \hline
      
\textbf{\begin{tabular}[c]{@{}c@{}}Simba~\cite{Xie2016_Simba}\\ (2016)\end{tabular}} &
      Spatial &
      Point &
      STR &
      \begin{tabular}[c]{@{}c@{}}R-Tree\\ (Multi-Level)\end{tabular} &
      \begin{tabular}[c]{@{}c@{}}Extended\\ Spark SQL\end{tabular} &
      \begin{tabular}[c]{@{}c@{}}Range, kNN,\\ kNN-Join,\\ Distance-Join,\end{tabular} \\ \hline
      
\textbf{\begin{tabular}[c]{@{}c@{}}LocationSpark~\cite{Tang2016_LocationSpark}\\ (2016)\end{tabular}} &
      Spatial &
      \begin{tabular}[c]{@{}c@{}}Point,\\ LineString,\\ Polygon,\\ Rectangle\end{tabular} &
      \begin{tabular}[c]{@{}c@{}}Uniform-Grid,\\ R-Tree,\\ Quad-Tree\\\end{tabular} &
      \begin{tabular}[c]{@{}c@{}}R-Tree\\Quad-Tree,\\IR-Tree\\(Multi-Level)\end{tabular} &
      N/A &
      \begin{tabular}[c]{@{}c@{}}Range, kNN,\\ Range-Join,\\ kNN-Join\end{tabular} \\ \hline
      
\textbf{\begin{tabular}[c]{@{}c@{}}SparkGIS~\cite{Furqan2017_SparkGIS}\\ (2017)\end{tabular}} &
      Spatial &
      \begin{tabular}[c]{@{}c@{}}Point,\\ LineString,\\ Polygon\end{tabular} &
      \begin{tabular}[c]{@{}c@{}}Fixed-Grid,\\ Binary-Space,\\ Quad-Tree,\\ Strip-based,\\ Hilbert-Curve,\\ STR\end{tabular} &
      \begin{tabular}[c]{@{}c@{}}R*-Tree\\ (Multi-Level)\end{tabular} &
      N/A &
      \begin{tabular}[c]{@{}c@{}}Range, kNN,\\ Join\end{tabular} \\ \hline
      
\textbf{\begin{tabular}[c]{@{}c@{}}Elcano~\cite{Englinus2018_Elcano}\\ (2018)\end{tabular}} &
      Spatial &
      \begin{tabular}[c]{@{}c@{}}Point,\\ LineString,\\ Polygon\end{tabular} &
      N/A &
      \begin{tabular}[c]{@{}c@{}}GeoHash,\\ R-Tree,\\ GeoHash + R-Tree\end{tabular} &
      \begin{tabular}[c]{@{}c@{}}Extended\\ Spark SQL\end{tabular} &
      Join \\ \hline
      
\textbf{\begin{tabular}[c]{@{}c@{}}STARK~\cite{Stefan2017_STARK}\\ (2017)\end{tabular}} &
      \begin{tabular}[c]{@{}c@{}}Spatial,\\Spatio-temporal\end{tabular} &
      \begin{tabular}[c]{@{}c@{}}STObject\\ (geo, time)\end{tabular} &
      \begin{tabular}[c]{@{}c@{}}Fixed-Grid,\\ Binary-Space\end{tabular} &
      \begin{tabular}[c]{@{}c@{}}R-Tree\\ (Live \& Persistent)\end{tabular} &
      Piglet &
      kNN, Join \\ \hline
      
\textbf{\begin{tabular}[c]{@{}c@{}}DiStRDF~\cite{Nikitopoulos2018_DiStRDF}\\ (2018)\end{tabular}} &
      Spatio-temporal &
      \begin{tabular}[c]{@{}c@{}}Point,\\Timestamp\end{tabular} &
      \begin{tabular}[c]{@{}c@{}}Range-Partition\end{tabular} &
      \begin{tabular}[c]{@{}c@{}}Hilbert Hash\\Z-order Hash\end{tabular} &
      SPARQL &
      Range, Join \\ \hline
      
\textbf{\begin{tabular}[c]{@{}c@{}}TrajSpark~\cite{Zhang2017_TrajSpark}\\ (2017)\end{tabular}} &
      \begin{tabular}[c]{@{}c@{}}Spatio-temporal\\ (Trajectory)\end{tabular} &
      \begin{tabular}[c]{@{}c@{}}Point,\\Timestamp\end{tabular} &
      \begin{tabular}[c]{@{}c@{}}Quad-Tree,\\KD-Tree \end{tabular} &
      \begin{tabular}[c]{@{}c@{}}Local: Hash,\\ Global: Multi-Level\\L1: Temporal\\L2: Spatial\\L3: B\textsuperscript{+}-Tree\end{tabular} &
      N/A &
      \begin{tabular}[c]{@{}c@{}}Single Object,\\Range,\\kNN \end{tabular} \\ \hline
      
\textbf{\begin{tabular}[c]{@{}c@{}}DITA~\cite{Zeyuan2018_DITA}\\(2018)\end{tabular}} &
      \begin{tabular}[c]{@{}c@{}}Spatio-temporal\\(Trajectory)\end{tabular} &
      \begin{tabular}[c]{@{}c@{}}Point,\\Timestamp\end{tabular} &
      STR &
      \begin{tabular}[c]{@{}c@{}}(Multi-Level)\\Global - R-Tree\\Local - Trie-like\end{tabular} &
      \begin{tabular}[c]{@{}c@{}}Extended\\ Spark SQL\end{tabular} &
      \begin{tabular}[c]{@{}c@{}}Similarity\\ Search and Join\end{tabular} \\ \hline

\textbf{\begin{tabular}[c]{@{}c@{}}UlTraMan~\cite{Ding2018_UITraMan}\\(2018)\end{tabular}} &
      \begin{tabular}[c]{@{}c@{}}Spatio-temporal\\(Trajectory)\end{tabular} &
      \begin{tabular}[c]{@{}c@{}}Point,\\TimeStamp\end{tabular} &
      STR &
      \begin{tabular}[c]{@{}c@{}}Two-Level\\(Global, Local)\\R-Tree\end{tabular} &
      N/A &
      \begin{tabular}[c]{@{}c@{}}ID, Range, \\ kNN,\end{tabular} \\ \hline
      
\textbf{\begin{tabular}[c]{@{}c@{}}Dragoon~\cite{Fang2021_Dragoon}\\(2021)\end{tabular}} &
      \begin{tabular}[c]{@{}c@{}}Spatio-temporal\\(trajectory)\\offline \& online\end{tabular} &
      \begin{tabular}[c]{@{}c@{}}Point,\\TimeStamp\end{tabular} &
      \begin{tabular}[c]{@{}c@{}}ID, Grid,\\STR, Time\end{tabular} &
      \begin{tabular}[c]{@{}c@{}}Two-Level\\(Global, Local)\\R-Tree\end{tabular} &
      N/A &
      \begin{tabular}[c]{@{}c@{}}ID, Range,\\kNN\end{tabular} \\ \hline
\end{tabular}}
\end{table}

GeoSpark~\cite{Yu2015_GeoSpark1, Yu2019_GeoSpark2} is a spatial extension of Spark, which extends Spark RDDs (Resilient Distributed Datasets)~\cite{Zaharia2012_Spark_RDD} to support spatial data types called Spatial RDD. It supports several spatial partitioning (Fixed-Grid, Voronoi Diagram, R-Tree, and Quad-Tree) and indexing (R-Tree and Quad-Tree) techniques to speed-up spatial queries (range, kNN, and join) on Spatial RDDs. Initially, it did not have any support for the SQL query~\cite{Yu2015_GeoSpark1}. Recently, GeoSpark has introduced an SQL API (SQL/MM-Part 3 Standard)~\cite{Yu2019_GeoSpark2} as a spatial extension of Spark SQL~\cite{Armbrust2015_SparkSQL}. SpatialSpark~\cite{You2015_SpatialSpark} can perform range queries and two kinds of spatial join queries (broadcast and partitioned) over geometric objects. Data can be partitioned using Fixed-Grid, Binary-Split, and Sort-Tile techniques and indexing using R-tree. However, it does not have support for SQL queries. Since SpatialSpark has implemented as a library on top of Spark instead of modifying the core of the framework, it may affect the query performance. Besides, both GeoSpark and SpatialSpark do not have any support for handling data and query skew. 

On the other hand, Simba~\cite{Xie2016_Simba} extends Spark SQL~\cite{Armbrust2015_SparkSQL} and DataFrame API to make spatial support for Spark. It improves the query performance by introducing multi-level (global and local) R-tree indexing on RDDs, and spatial-aware (logical and cost-based) query planning. Moreover, STR partitioner~\cite{Leutenegger1997_STR} mitigates the data partitioning skew significantly due to its consideration of in-memory partition size, data locality, and load balancing. However, Simba only supports spatial operations (range, kNN, distance-join, kNN-join) over point and rectangle objects. LocationSpark~\cite{Tang2016_LocationSpark, Tang2019_LocationSpark} was developed as a spatial library (like SpatialSpark) over Spark. It stores spatial data as a key-value pair, where the key can be any geometric object (points, lines, polygons), and the values can be any user-specified text. Like Simba, it also contains an efficient cost model and a query execution planner to deal with data partitioning and query skew. Similarly, it supports multi-level indexing, where the global index (grid, region Quad-tree) partitions the data across a cluster of nodes and a local index (R-tree, Quad-tree variant, IR-tree) for indexing data on each node. Moreover, it introduced a spatial bloom filter to reduce the communication cost of the global index. LocationSpark only keeps frequently accessed data in memory which reduces the chances of an overflow. It supports a number of spatial operations (such as range, kNN, range-join, and kNN-join) and a few spatial analysis functions (such as clustering, skyline computation, and spatio-textual topic summarization). However, it does not have support for SQL-like queries. SparkGIS~\cite{Furqan2017_SparkGIS} adopts Spark for processing spatial queries (kNN, join). It supports several dynamic partitioning algorithms (Fixed-Grid, Binary-Space, Quad-Tree, Strip-based, Hilbert-Curve, and STR), which mitigates the data distribution skew across the cluster. Like Simba and LocationSpark, it also incorporates multi-level (global and local) R*-tree indexing, which can be pre-generated or on-demand local in-memory indexing. Like LocationSpark, it also keeps data in-memory as much as possible to avoid running out of memory. 

However, these five spatial data processing solutions are not fully compliant with the ISO standard and OGC specifications. Elcano~\cite{Englinus2018_Elcano} implements ISO and OGC-compliant 2D geometry data types, spatial functions, and operators based on Spark SQL~\cite{Armbrust2015_SparkSQL}. It supports three indexation methods, GeoHash, R-Tree, and a combination of both (hybrid)~\cite{Englinus2019_Elcano2}. This paper reports the spatial join performance of Elcano over SpatialSpark and PostgreSQL/PostGIS. However, this paper did not include any information regarding data partitioning, and the process of spatial query execution of Elcano.

All these Spark-based systems discussed above are only for spatial data processing. STARK~\cite{Stefan2017_STARK} is a spatio-temporal data processing system. It integrates spatial and temporal data types, operators, and predicates to Spark RDD's. It supports fixed grid and cost-based binary space partitioning to distributed data across the nodes of a cluster. It allows two modes of indexing, where the live index is built for each partition during query execution, and persistent indexing allows to create and save indexed RDD into disk or HDFS for future use. STARK supports queries on unindexed data as well. In addition to spatial join and kNN query, it supports DBSCAN clustering. STARK also extends Pig Latin for declarative spatio-temporal queries, called Piglet. Though STARK has support for temporal features, the only reported  evaluation results are for spatial operations. DiStRDF~\cite{Nikitopoulos2018_DiStRDF} is a distributed Spatio-temporal RDF~\cite{url:RDF} data processing system based on Spark. It consists of two layers, where the storage layer is responsible for storing encoded RDF triples into HDFS and dictionary of mapping values into the Redis in-memory key-value store~\cite{url:redis}. The query processing layer is based on the Spark query engine, which is responsible for parsing, planning, and executing SPARQL~\cite{url:SPARQL} queries. Here, Apache Jena~\cite{url:Jena} is used as a query parser. It uses spatio-temporal range partitioning to distribute 1D encoded RDF triples. It also supports Hilbert and Z-order hashing for indexing RDF triples. However, it only supports  spatio-temporal point data.

As it is mentioned in Section~\ref{sub-sec:hadoop-systems}, trajectory data is quite different from discrete spatio-temporal point data. Therefore, the performance of processing trajectory data using STARK and DiStRFD will not be effective. Also, these systems work well for historical static data and require re-partitioning the whole dataset when the dataset has changed or updated. TrajSpark~\cite{Zhang2017_TrajSpark} always keeps the global index in the main memory and updates the global index when new data arrives using the time-decay model by partitioning only a batch of new data. TrajSpark also stores the updated the global index into a disk to protect it from any future system failures. In TrajSpark, first, the raw trajectory points (RDD) are partitioned based on data locality, load balancing, and size of the partition using a Quad-Tree/KD-Tree strategy. Then the local hash index is added to each partition, which creates an IndexTRDD. Finally, a multi-level hybrid global index (level1 - temporal, level2 - spatial, level3 - B\textsuperscript{+}-Tree) is built for each partition. One can perform single object-based, range, and kNN queries on trajectory data using TrajSpark. However, it does not have any support for  SQL-like queries. Zeyuan et al.~\cite{Zeyuan2018_DITA} developed DITA, a distributed in-memory trajectory analytics system, where one can use both SQL and Dataframe API for trajectory analysis. DITA adopts the STR~\cite{Leutenegger1997_STR} partitioning strategy to create balanced partitions of trajectory points. Like TrajSpark, it also uses multi-level (global - R-Tree and local - Trie-like) indexing to expedite the query performance. Besides, DITA has developed a cost model to reduce inter-worker transmission costs and to balance the workload. Like DITA, UITraMan~\cite{Ding2018_UITraMan} also uses STR for partitioning trajectory data, but it has adopted R-Tree for both local and global indexing. Along with global indexing, UITraMan also maintains a meta table to store information related to moving objects and partitions in order to improve the efficiency of trajectory data processing in Spark. Unlike other systems, UITraMan incorporates a data processing pipeline that includes data loading, preprocessing, extraction, and analysis. However, the on-heap data caching in Spark induces GC (garbage collector) overhead, and the performance of Spark-based systems is affected by this overhead. Therefore, UITraMan has added an off-heap key-value store Chronicle Map~\cite{url:chronicle-map} into the block manager of Spark. Chronicle Map always keeps data in an off-heap cache, which reduces GC overhead and ensures data persistence on run-time. UITraMan supports ID, range, and kNN queries on trajectory data.

Among TrajSpark, DITA, and UITraMan, only TrajSpark alleviates the overhead of re-partitioning the whole dataset when a new batch of dataset arrives. Thus, TrajSpark achieves near real-time trajectory processing capability, but it is not a system developed for processing real-time trajectory streams. Besides, this new batch of data is loaded as RDDs in Spark, which are immutable, and any updates on RDD create a new RDD, which is costly. Dragoon~\cite{Fang2021_Dragoon} is a hybrid trajectory analytics system for processing both historical (offline) and streaming (online) trajectories. The offline module of Dragoon is similar to UITraMan, but Dragoon has utilized Chronicle Map in such a way that it works for both historical and streaming trajectories. In addition, a mutable RDD (mRDD) model is designed so that data can be updated later, which is key to the hybrid storage of Dragoon. In Dragoon, data partitioning (ID, spatial: Grid, STR, and temporal), indexing (two-level: R-Tree), and trajectory queries (ID, range, and kNN) are developed for both offline and online modules. Moreover, the hybrid data processing pipeline provides support for both historical and streaming trajectories.

Other than these systems, GeoMesa~\cite{url:geomesa} has recently added support for Spark. However, all these systems are for processing vector spatial and spatio-temporal data. None of these systems has support for raster data. GeoTrellis~\cite{url:GeoTrellis} is a Scala library that enables Spark to process spatial raster data. It also has limited support for vector data. It can store into and query raster data from HDFS, S3, Accumulo, Cassandra, and HBase. 


\subsection{NoSQL-based Big Spatio-temporal Infrastructures}
\label{sub-sec:nosql-systems}
 A number of big spatio-temporal data processing systems have been developed by using the power of NoSQL databases in the last couple of years, such as MD-HBase~\cite{Nishimura2011_MD-HBase}, Distributed SECONDO~\cite{Ralf2015_DistSECONDO}, GeoMesa~\cite{url:geomesa, Fox2013_GeoMesa, Hughes2015_GeoMesa}, BBoxDB~\cite{Nidzwetzki2018_BBoxDB, Nidzwetzki2019_BBoxDB_Demo}, THBase~\cite{Qin2019_THBase}, TrajMesa~\cite{Li2020_TrajMesa}, and JUST~\cite{Li2020_JUST}. The detailed feature matrix of these systems is presented in Table~\ref{tab:nosql_based_systems}.

\begin{table}[ht!]
\centering
\caption{NoSQL-based Spatio-temporal Systems}
\label{tab:nosql_based_systems}
\resizebox{\textwidth}{!}{%
\begin{tabular}{|c|c|c|c|c|c|c|c|}
\hline
  &
  \textbf{System Type} &
  \textbf{\begin{tabular}[c]{@{}c@{}}Underlying\\ NoSQL System\end{tabular}} &
  \textbf{Data Types} &
  \textbf{Partitioning} &
  \textbf{Indexing} &
  \textbf{\begin{tabular}[c]{@{}c@{}}Query\\ Language\end{tabular}} &
  \textbf{\begin{tabular}[c]{@{}c@{}}Supported\\ Queries\end{tabular}} \\ \hline
  
\textbf{\begin{tabular}[c]{@{}c@{}}MD-HBase~\cite{Nishimura2011_MD-HBase}\\ (2011)\end{tabular}} &
  Spatio-temporal &
  HBase &
  \begin{tabular}[c]{@{}c@{}}Point,\\ Timestamp\end{tabular} &
  Range-Partition &
  \begin{tabular}[c]{@{}c@{}}Quad-Tree,\\ KD-Tree\end{tabular} &
  N/A &
  Range, kNN \\ \hline
  
\textbf{\begin{tabular}[c]{@{}c@{}}GeoMesa~\cite{url:geomesa}\\ (2013)\end{tabular}} &
  Spatio-temporal &
  Accumulo &
  \begin{tabular}[c]{@{}c@{}}Point,\\ LineString,\\ Polygon,\\ Timestamp\end{tabular} &
  \begin{tabular}[c]{@{}c@{}}Spatial,\\ Temporal,\\ Attribute\end{tabular} &
  \begin{tabular}[c]{@{}c@{}}Z2 and XZ2,\\ Z3 and XZ3,.\\ id and attr\end{tabular} &
  CQL &
  Range \\ \hline
  
\textbf{\begin{tabular}[c]{@{}c@{}}Distributed\\SECONDO~\cite{Ralf2015_DistSECONDO}\\ (2015)\end{tabular}} &
  \begin{tabular}[c]{@{}c@{}}Spatio-temporal\\ (Trajectory)\end{tabular} &
  Cassandra &
  \begin{tabular}[c]{@{}c@{}}Point, LineString,\\ Regions, Instant,\\ Period, Periods, \\ Interval\end{tabular} &
  3D Grid &
  \begin{tabular}[c]{@{}c@{}}R-tree \\ (MMR-tree)\end{tabular} &
  \begin{tabular}[c]{@{}c@{}}SQL-Like,\\ Executable\end{tabular} &
  Join \\ \hline
  
\textbf{\begin{tabular}[c]{@{}c@{}}BBoxDB~\cite{Nidzwetzki2018_BBoxDB}\\(2018)\end{tabular}} &
  Spatial &
  \begin{tabular}[c]{@{}c@{}}Key-BBox-Value\\ Store\end{tabular} &
  \begin{tabular}[c]{@{}c@{}}Point,\\ LineString,\\ Polygon\end{tabular} &
  \begin{tabular}[c]{@{}c@{}}Grid,\\ KD-Tree,\\ Quad-Tree\end{tabular} &
  \begin{tabular}[c]{@{}c@{}}Two-Level\\ Local:  R-Tree\\ Global:  KD-Tree\end{tabular} &
  N/A &
  Join \\ \hline

\textbf{\begin{tabular}[c]{@{}c@{}}THBase~\cite{Qin2019_THBase}\\(2019)\end{tabular}} &
  \begin{tabular}[c]{@{}c@{}}Spatio-temporal\\(Trajectory)\end{tabular} &
  \multicolumn{1}{c|}{HBase} &
  \begin{tabular}[c]{@{}c@{}}Point,\\Timestamp\end{tabular} &
  \begin{tabular}[c]{@{}c@{}}MO-based\\Model\end{tabular} &
  \begin{tabular}[c]{@{}c@{}}Two-Level\\L1: Time Index\\L2: Multi-level Grid\end{tabular} &
  N/A &
  \begin{tabular}[c]{@{}c@{}}Single-Object,\\Range, kNN\end{tabular} \\ \hline
  
\textbf{\begin{tabular}[c]{@{}c@{}}JUST~\cite{Li2020_JUST}\\(2020)\end{tabular}} &
  \begin{tabular}[c]{@{}c@{}}Spatial,\\ Spatio-temporal\end{tabular} &
  HBase &
  \begin{tabular}[c]{@{}c@{}}Geom, \\ Timestamp,\\ ST\_Series,\\ T\_Series\end{tabular} &
  N/A &
  \begin{tabular}[c]{@{}c@{}}Z2 and XZ2,\\ Z3 and XZ3,\\ Z2T and XZ2T\end{tabular} &
  JustQL &
  Range, kNN \\ \hline
  
\textbf{\begin{tabular}[c]{@{}c@{}}TrajMesa~\cite{Li2020_TrajMesa}\\(2020)\end{tabular}} &
  \begin{tabular}[c]{@{}c@{}}Spatio-temporal\\ (Trajectory)\end{tabular} &
  GeoMesa &
  \begin{tabular}[c]{@{}c@{}}Point,\\ Timestamp\end{tabular} &
  N/A &
  XZT and XZ2\textsuperscript{+} &
  SQL-Like &
  \begin{tabular}[c]{@{}c@{}}ID-Temporal\\ Range, kNN,\\ Similarity\end{tabular} \\ \hline
  
\end{tabular}%
}
\end{table}

MD-HBase~\cite{Nishimura2011_MD-HBase} extends HBase~\cite{url:hbase} to support spatio-temporal queries (range and kNN). It applies linearization (e.g., Z-Ordering) to transform multi-dimensional locations data (id, lat, lon, time) into 1D space for efficient indexing. A multi-dimensional index structure (Quad-tree, KD-tree) is layered on top of a range partitioned key-value store which allows real-time processing of range and kNN queries. MD-HBase achieves high insertion throughput, which is important for location-based applications. GeoMesa~\cite{url:geomesa} is a spatio-temporal data processing system built on  top of NoSQL databases that provides efficient storage and querying capabilities. It was implemented based on the distributed key-value store Accumulo~\cite{Fox2013_GeoMesa}. Like MD-HBase, GeoMesa also linearizes the keyspace by transforming multi-dimensional data (location, timestamp) into 1D keys using space-filling curves. It creates a spatio-temporal index using GeoHash and timestamps. Later, GeoMesa~\cite{url:geomesa, Hughes2015_GeoMesa} has added support for HBase, Google BigTable, Cassandra, Kafka, and Spark. At present, GeoMesa supports a set of indexing techniques, such as spatial (Z2 and XZ2), spatio-temporal (Z3, XZ3), ID index, and attribute index. Traditional key-value stores with multi-dimensional support could be expensive to store and query non-point spatial data such as polygons or lines. Jan et al.~\cite{Nidzwetzki2018_BBoxDB} have proposed a distributed and scalable key-bounding-box-value store for multi-dimensional data called BBoxDB. Unlike traditional key-value stores, BBoxDB stores each value with an n-dimensional axis-parallel bounding box, which defines the location of the value in space. It uses space partitioning (Grid, KD-Tree, Quad-Tree) and multi-level indexing (global: KD-Tree, local: R-Tree) to store and organize the data across the cluster of nodes. However, BBoxDB only supports spatial-join queries, which can be executed locally on co-partitioned data.

Distributed SECONDO~\cite{Ralf2015_DistSECONDO, Ralf2017_DistSECONDO} is a general-purpose DBMS, which can process relational, spatial, and spatio-temporal (including trajectory) data. It integrates the highly scalable and available key-value store Apache Cassandra and moving objects database SECONDO~\cite{Guting2010_Secondo}, where Cassandra is used as distributed data storage and the SECONDO as a query processing engine. Previously, SECONDO was integrated with Hadoop in Parallel SECONDO, but suffered due to centralized management. Besides, Parallel SECONDO~\cite{Ralf2015_PSecondo} does not support high update rates. Distributed SECONDO achives high update rates by integrating Cassandra. In addition, it supports both SQL-like and executable query. JUST (JD Urban Spatio-Temporal)~\cite{Li2020_JUST} incorporates the power of HBase~\cite{url:hbase}, GeoMesa~\cite{url:geomesa, Fox2013_GeoMesa, Hughes2015_GeoMesa}, and Spark~\cite{url:spark} into one system to stores, manages, and processes spatio-temporal data. It adopts the NoSQL store HBase as an underlying storage structure, GeoMesa as an indexing tool, and Spark as a query execution engine. Along with indexing strategies of GeoMesa, JUST introduces two indexing techniques, Z2T and XZ2T and efficient compression mechanism to expedites the query performance significantly. A SQL-like query language JustQL was also developed in JUST from scratch. Unlike in-memory systems, JUST only loads the necessary data into memory. Hence, the nodes of a cluster are not required to have much memory. Though NoSQL-based systems (mentioned before) have achieved high update rates, their performance is hindered by disk latency. JUST has improved query efficiency by utilizing the main memory. 

TrajMesa~\cite{Li2020_TrajMesa} has adopted GeoMesa to develop a trajectory storage engine, where a horizontal storage schema (H-Store) is proposed for efficient trajectory data management. Instead of storing each point of a trajectory as a separate entry in a key-value store (V-Store), H-Store allows storing an entire trajectory in one-row with compression. Hence, in addition to reducing the storage size significantly, H-Store also improves the query efficiency by reducing disk I/O in TrajMesa. To perform a set of SQL-like queries (ID-Temporal, Range, kNN, and Similarity) on trajectory data efficiently, TrajMesa introduced ID temporal, XZT, and extended XZ2 indexing of GeoMesa as spatial range indexing, XZ2\textsuperscript{+}. Most importantly, TrajMesa incorporates a module for trajectory preprocessing~\cite{Ruan2018_CloudTP} containing functions for noise filtering, segmentation, stay point detection, map matching, and other statistical analysis. On the other hand, if we split and store the trajectories of the same moving object (MO) into different partitions on different nodes of a cluster, the query processing efficiency of trajectory processing systems~\cite{Zhang2017_TrajSpark, Ding2018_UITraMan} will be impacted negatively. To address this issue, THBase~\cite{Qin2019_THBase} has proposed a segment-based data model and a MO-based partition model for efficient trajectory storage management in HBase. THBase consists of three modules, T-table, L-index, and a query processing module. T-table is a container of trajectory data in which adopted the MO-based partitioning model. Whereas, L-index is a local spatio-temporal index structure that consists of two levels (level1: time index, level2: Quad-tree based multi-level grid). Finally, the query processing module supports single-object, spatio-temporal range, and kNN query. 


\subsection{Python Libraries as Big Spatio-temporal Infrastructures}
\label{sub-sec:python_big_library}

Python is one of the most popular data analytics platforms today. The PyData stack is rich in terms of supported libraries, but most of these libraries are developed to execute on a single CPU core and to process data that fits in main memory. Therefore, Python does not scale well for processing big data. One can utilize PySpark~\cite{Team2020_PySpark} to process data using Spark~\cite{url:spark} on a cluster of nodes. At present, Spark is the most popular distributed in-memory general-purpose data processing engine that supports a set of essential libraries (such as Spark SQL, Streaming, MLlib, and GraphX). Spark has a rich support community, and users can write code in many languages (such as Java, Scala, Python, and R). However, PySpark is added as an extra layer on top of Spark, and therefore, when Python code is executed using PySpark, the code is first compiled into Java code and then run on JVM. Thus, PySpark adds an extra overhead in computation. On the contrary, DASK~\cite{Team2016_Dask} is a Python library for parallel and distributed computing that scales Python natively. DASK  not only scales Python across distributed nodes of a cluster, but it also parallelizes a task in a single node by utilizing multiple CPU cores. Therefore, those who process data using Python on a personal computer regularly can easily speedup computation using DASK. 

DASK is a task-graph based platform that consists of two parts. At high-level, DASK parallelizes the Python ecosystem by extending existing libraries, such as Pandas (DASK DataFrame), Numpy (DASK Array), Scikit-Learn (DASK-ML), and other libraries. Whereas, at low-level, DASK provides dynamic task schedulers like Airflow or Luigi to enable advanced workloads. Here, DASK libraries (Array, DataFrame) produce task graphs and then DASK schedulers execute task graphs parallelly either in a single node or in  distributed nodes of a cluster.  Python also supports a rich set of libraries (e.g., GeoPandas) for processing spatial, spatio-temporal, and trajectory data. A detailed review of these libraries is presented in Section~\ref{sub-sec:st_python}. However, these libraries are also slow and not scalable to process big spatial data like other Python libraries. DASK does not have a native parallel module for processing big spatial data. One can utilize DASK DataFrame or low-level capabilities of DASK with existing spatial libraries to process big spatial data. Previously, developers have tried to improve the performance of GeoPandas through Cython~\cite{behnel2011_cython}, which allows GeoPandas to access the GEOS library directly.  However, Cythonizing only utilized one core of a single node effectively~\cite{url:FastGeoPandas}. Hence, the main challenge is to use multiple CPU cores or distributed nodes of a cluster. Recently, DASK-GeoPandas~\cite{url:dask-geopandas} was developed to parallelize GeoPandas with DASK, which organizes many GeoPandas dataframes like DASK-DataFrame. However, this is an experimental project and currently partitions dataframes by rows. Thus, spatial operations will not return the correct results in a distributed environment without using spatial partitioning.  XArray~\cite{hoyer2017_xarray} is a Python package for labeled multi-dimensional arrays which is efficient for processing scientific datasets (e.g., netCDF, GeoTiff). Since XArray is tightly integrated with DASK for parallel computation, we can utilize XArray for processing big raster spatial data. A few other projects, such as dask-geomodeling~\cite{url:dask-geomodeling} and dask-rasterio~\cite{url:dask-rasterio}, also provide support for spatial raster data. Moreover, dask-geomodeling has a module for spatial vector data. 

As mentioned before,  Hadoop is an efficient framework for big data processing but hindered by the I/O bottleneck. In this context,  as Spark keeps data always in-memory and does not require to write intermediate results back to disk,  it became an efficient and popular data processing framework. Currently, CPU-based in-memory systems like Spark suffer due to the  bottleneck of processing complex workloads (e.g., deep learning on massive datasets), and this bottleneck is due to the CPU itself. Compared to  CPUs with a few cores and lots of cache memory, GPUs are composed of hundreds of cores, high bandwidth memory (up to TB/s), and high-speed hardware interconnections (e.g., bidirectional GPU to GPU bandwidth up to 300 GB/s). Also, GPUs can scale up to 16x in a single node~\cite{url:Talk_RAPIDS}. RAPIDS~\cite{2018_RAPIDS} is a collection of libraries and APIs that bring the power of GPUs for processing big data in Python. RAPIDS supports a wide range of libraries for Python developers, such as analytics (cuDF, cuIO), machine learning (cuML), graph analytics (cuGraph), deep learning (PyTorch, TensorFlow, MxNet), spatial analytics (cuSpatial), visualization (cuxFilter, pyViz, plotly), and other libraries. These libraries are literally replicated versions of the existing python libraries. Therefore, Python developers can utilize the processing capability of GPUs without knowing low-level CUDA implementations. Using RAPIDS, we can achieve vertical scalability easily. Since RAPIDS integrates DASK, horizontal scalability can be achieved through RAPIDS and DASK using OpenUCX on a single node as well as in distributed nodes of a cluster.

cuSpatial is a spatial module of RAPIDS, which is still in early-stages of development, but growing rapidly. Dependent on the type of operations, it is possible to achieve significant (up to 1000x) performance improvement in RAPIDS when compared to CPU spatial libraries like GDAL. cuSpatial integrates with RAPIDS dataframe cuDF to use GPUs massive parallelism and high memory bandwidth for performing spatial operations. Developers can also use cuSpatial and cuGraph together for spatial and spatio-temporal analytics. cuSpatial is implementing spatial features in four layers that include geometry types, spatial operations, indexing, and querying. The current version of cuSpatial (v0.16) can model some basic geometry types such as points, polylines, polygons, and shape primitives. It supports Quad-Tree indexing for performing various spatial operations, such as point-in-polygon, Haversine distance (distance between points), and Hausdorff distance (distance between trajectories). Currently, it supports spatial window and nearest polyline queries. Since cuSpatial is working seamlessly with cuDF, users can use various data formats, such as CSV, Parquet, Shapefiles, JSON, and more.

In terms of support, Spark lacks data visualization and deep learning libraries. Whereas, both DASK and RAPIDS have support for data visualization and deep learning libraries (PyTorch, TensorFlow, Keras). At present, there is no SQL query support in DASK. However, one can run SQL queries on both Spark (Spark SQL) and RAPIDS (blazingSQL). Though all of them have support for machine learning libraries,  DASK (Dask-ML) and RAPIDS (cuML) libraries are more popular compared to Spark (MLib). 


\subsection{Other Big Spatio-temporal Infrastructures}
\label{sub-sec:other-systems}
A number of distributed systems were developed by extending existing infrastructures other than Hadoop, Spark, and NoSQL-databases for processing spatial data. Sphinx~\cite{Eldawy2015_Sphinx, Eldawy2017_Sphinx} is a highly-scalable distributed spatial data processing system that extends the core of Apache Impala~\cite{url:impala, Kornacker2015_Impala}, which is a SQL engine over Hadoop. Sphinx has adopted the ANSI-standard SQL interface and built spatial support in four layers of Impala. It introduced spatial data types, OGC-compliant spatial predicates and functions, and commands to create spatial indexes in the query parser layer. Two-level index is added in the storage layer, where the global index (R-tree/R+-tree) partitions the data into blocks, and local indexes (R-tree) arrange records in each block of HDFS. Finally, Sphinx modifies the query planner and the executor layer to add support for spatial joins and range queries. Since SQL-like queries of most big spatial data processing systems are not ANSI-standard and not efficient as spatial RDBMS, Sphinx achieves good performance over systems like SpatialHadoop. However, one can run a query on existing data only, as data updates are not allowed in HDFS.  AsterixDB~\cite{Alsubaiee2014_AsterixDB, Alsubaiee2014_AsterixDB_Storage, Alamoudi2015_AsterixDB} is a full-fledged big data management system, which incorporates LSM-based data storage and a set of indexing techniques including B\textsuperscript{+}-tree and R-tree. It supports a complete query language, AQL, that uses Hyracks~\cite{Borkar2011_Hyracks} as a query execution engine. A rich set of built-in data types, including spatial and temporal data, allows users to perform spatial, temporal, and spatio-temporal queries. Currently, it also supports the SQL++ query language, which is very similar to SQL, but for semi-structured data (e.g., JSON).


Like Spark, Apache Ignite~\cite{url:ignite} is also a scalable and fault-tolerant distributed in-memory big data processing platform, but  has limited spatial supports. Its geospatial module supports geometry data types (point, line, and polygon), a limited form of querying on geometry data (intersection operation), and spatial indexing R-tree from the H2 database~\cite{url:H2DB}. The main limitation of Ignite is that it does not support any spatial partitioning technique to distribute data across the clusters. Therefore, the result returned from Ignite for any spatial query is not accurate. Alam et al.\cite{Mahbub2018_BigSpatial_PStudy} introduced $SpatialIgnite$~ as extended spatial support for Ignite. They have added a spatial library containing all the OGC-compliant spatial predicates and analysis functions and introduced two spatial data partitioning techniques (fixed grid and Niharika~\cite{Ray2011_Niharika}) into $SpatialIgnite$. According to the reported evaluation results, $SpatialIgnite$ outperforms GeoSpark~\cite{Yu2015_GeoSpark1} for executing spatial join and range queries.

\begin{table}[ht!]
\centering
\caption{Other Big Spatio-temporal Infrastructures}
\label{tab:other_big_systems}
\resizebox{\textwidth}{!}{%
\begin{tabular}{|c|c|c|c|c|c|c|c|}
\hline
\textbf{} &
  \textbf{System Type} &
  \textbf{\begin{tabular}[c]{@{}c@{}}Underlying\\ System/Architecture\end{tabular}} &
  \textbf{\begin{tabular}[c]{@{}c@{}}Geometry\\ Types\end{tabular}} &
  \textbf{Partitioning} &
  \textbf{Indexing} &
  \textbf{\begin{tabular}[c]{@{}c@{}}Query\\ Language\end{tabular}} &
  \textbf{\begin{tabular}[c]{@{}c@{}}Supported\\ Queries\end{tabular}} \\ \hline
  
\textbf{\begin{tabular}[c]{@{}c@{}}RasDaMan~\cite{Baumann1997_RasDaMan}\\ (1997)\end{tabular}} &
  Spatial Raster &
  Array DB &
  N/A &
  N/A &
  N/A &
  RasQL &
  Range \\ \hline
  
\textbf{\begin{tabular}[c]{@{}c@{}}SciDB~\cite{Cudre2009_SciDB, Brown2010_SciDB}\\ (2009)\end{tabular}} &
  Spatial Raster &
  Array DB &
  N/A &
  N/A &
  KD-Tree &
  AFL, AQL &
  Range, kNN \\ \hline
  
\textbf{\begin{tabular}[c]{@{}c@{}}DISTIL~\cite{Maria2018_DISTIL, Maria2019_DISTIL}\\ (2018)\end{tabular}} &
  Spatio-temporal &
  APGAS\textsuperscript{1} &
  \begin{tabular}[c]{@{}c@{}}Point (id,\\ oid, lat, lon,\\ timestamp,\\ direction,\\ speed)\end{tabular} &
  \begin{tabular}[c]{@{}c@{}}RRR\textsuperscript{2},\\ MDR\textsuperscript{3}\end{tabular} &
  \begin{tabular}[c]{@{}c@{}}Multi-Level\\ L1: Quad-Tree\\ L2: Spatial\\ L3: Temporal\end{tabular} &
  N/A &
  Range, kNN \\ \hline \hline
  
\textbf{\begin{tabular}[c]{@{}c@{}}AsterixDB~\cite{Alsubaiee2014_AsterixDB}\\ (2014)\end{tabular}} &
  \begin{tabular}[c]{@{}c@{}}Spatial,\\ Spatio-temporal\end{tabular} &
  Hyracks~\cite{Borkar2011_Hyracks} &
  \begin{tabular}[c]{@{}c@{}}Point, Line,\\ Polygon, Circle, \\ Rectangle, Date, \\ Time, Interval, \\ Duration, etc.\end{tabular} &
  Hash &
  \begin{tabular}[c]{@{}c@{}}B+-Tree,\\ R-tree\end{tabular} &
  AQL, SQL++ &
  Range, Join \\ \hline
  
\textbf{\begin{tabular}[c]{@{}c@{}}Sphinx~\cite{Eldawy2015_Sphinx}\\ (2015)\end{tabular}} &
  Spatial &
  Apache Impala &
  \begin{tabular}[c]{@{}c@{}}Point, \\ LineString,\\ Polygon,\\ Collections\end{tabular} &
  STR &
  \begin{tabular}[c]{@{}c@{}}Two-Level\\ Global: R/R+-Tree\\ Local: R-Tree\end{tabular} &
  \begin{tabular}[c]{@{}c@{}}SQL\\ (ANSI-Standard)\end{tabular} &
  Range, Join \\ \hline
  
\textbf{\begin{tabular}[c]{@{}c@{}}SpatialIgnite~\cite{Mahbub2018_BigSpatial_PStudy}\\ (2018)\end{tabular}} &
  Spatial &
  Apache Ignite &
  \begin{tabular}[c]{@{}c@{}}Point,\\ LineString,\\ Polygon\end{tabular} &
  \begin{tabular}[c]{@{}c@{}}Grid,\\ Niharika~\cite{Ray2011_Niharika}\end{tabular} &
  R-Tree &
  \begin{tabular}[c]{@{}c@{}}Distributed SQL\\ (ANSI-Standard)\end{tabular} &
  Range, Join \\ \hline \hline
  
\textbf{\begin{tabular}[c]{@{}c@{}}Tornado~\cite{Mahmood2015_Tornado}\\ (2015)\end{tabular}} &
  \begin{tabular}[c]{@{}c@{}}Spatio-textual\\ Stream\end{tabular} &
  Apache Storm &
  \begin{tabular}[c]{@{}c@{}}\{srcid, oid, \\ (x, y), t, text\}\end{tabular} &
  A-Grid &
  \begin{tabular}[c]{@{}c@{}}Adaptive Indexing\\ Global: Spatial (A-Grid)\\ Local: Spatio-textual\\ (KD-Tree)\end{tabular} &
  \begin{tabular}[c]{@{}c@{}}Atlas\\ (SQL-Like)\end{tabular} &
  \begin{tabular}[c]{@{}c@{}}Snapshot,\\ Continuous\\ (Range, kNN,\\ Join)\end{tabular} \\ \hline
  
\textbf{\begin{tabular}[c]{@{}c@{}}SSTD~\cite{Chen2020_SSTD}\\ (2020)\end{tabular}} &
  \begin{tabular}[c]{@{}c@{}}Spatio-textual\\ Stream\end{tabular} &
  Apache Storm &
  Point &
  \begin{tabular}[c]{@{}c@{}}QT-tree\\ (Spatial,\\ Textual)\end{tabular} &
  \begin{tabular}[c]{@{}c@{}}Global: QT-tree\\ Local: Object, Query\end{tabular} &
  N/A &
  \begin{tabular}[c]{@{}c@{}}Snapshot,\\ Continuous\\ (Range, kNN,\\ Top-k)\end{tabular} \\ \hline
  
\textbf{\begin{tabular}[c]{@{}c@{}}GeoFlink~\cite{Shaikh2020_GeoFlink}\\ (2020)\end{tabular}} &
  Spatial Stream &
  Apache Flink &
  Point &
  Grid &
  Grid-based &
  N/A &
  \begin{tabular}[c]{@{}c@{}}Continuous\\ (Range, kNN,\\ Join)\end{tabular} \\ \hline
\end{tabular}%
}
\raggedright \tiny \textsuperscript{1} APGAS - Asynchronous Partitioned Global Address Space \\
\raggedright \tiny \textsuperscript{2} RRR - Row-wise Round-Robin Partitioning \\
\raggedright \tiny \textsuperscript{3} MDR - Multi-dimensional Range Partitioning
\end{table}

There are a few spatial database systems that have been developed from scratch, such as SciDB~\cite{Cudre2009_SciDB, Brown2010_SciDB, Stonebraker2011_SciDB, Becla2013_SciDB}, RasDaMan~\cite{Baumann1997_RasDaMan, Baumann1998_RasDaMan, Baumann1999_RasDaMan}, and DISTIL~\cite{Maria2018_DISTIL, Maria2019_DISTIL}. Both RasDaMan and SciDB are specialized database systems developed from scratch for scientific computing. These systems are implemented using a multi-dimensional array data model and efficient for processing spatial raster data. Besides, these systems support SQL-like queries. DISTIL is a scalable spatio-temporal in-memory system that is implemented based on APGAS (Asynchronous Partitioned Global Address Space) programming model. Its efficient data partitioning and distributed multi-level spatio-temporal indexing expedites the performance of range and kNN queries. DISTIL achieves a high rate of updates by incorporating LSM-Tree~\cite{ONeil1996_LSMTree} based key-value store LevelDB~\cite{url:LevelDBJ} (developed by Google) as a local data store in each node of a cluster. Besides, the data in the local store is periodically synchronized with the distributed persistent global store, HDFS. At present, DISTIL does not have any support for SQL-like queries. However, data processing systems built from scratch for a specific purpose, such as spatial data processing, can achieve good performance. Whereas, it is always challenging to develop a full-fledged system from scratch, and also, it is hard to use them as a general-purpose system~\cite{Eldawy2015_BigSpatial_Survey}. Besides,  the code base of these systems is frequently immature and difficult to extend.

The big spatial data processing systems, which are developed based on Hadoop, Spark, or other platforms, can only store and process  historical static spatial data. However, a wide range of location-based services require real-time processing of spatial data streams. Also, the indexing of these systems does not support high update rates to adjust with newly arriving streams. Besides, these services demand dynamic workload distribution. GeoFlink~\cite{Shaikh2020_GeoFlink} extends Apache Flink to add support for processing spatial data streams. It has introduced grid-based dynamic indexing to perform continuous queries (range, kNN, and join). Both Tornado~\cite{Mahmood2015_Tornado, Mahmood2017_Tornado} and SSTD~\cite{Chen2020_SSTD} have extended Apache Storm for processing spatio-textual (e.g., geo-tagged tweets) data streams. Tornado has added a two-level (global spatial and local spatio-textual) adaptive indexing layer for dynamically distributing data and query workloads. However, global indexing (A-Grid) and the cost model for load balancing of Tornado works well with continuous queries, not snapshot queries. Hence, SSTD has introduced QT-tree (a Quad-tree variant) global indexing and a set of local indexing for both continuous and snapshot queries (range, kNN, and top-k). Among them, only Tornado supports SQL-like (Atlas) queries with a map-based interface. 

Currently, most major commercial data management systems have some form of spatial support. For example, Google BigQuery~\cite{url:bigquery} has a GIS module (BigQuery GIS) to perform ANSI-standard SQL queries on large spatial datasets. Users can also visualize BigQuery results using BigQuery Geo Viz and Google Earth Engine. Google Earth Engine~\cite{Noel2017_GoogleEE} is itself a cloud-based platform for the analysis of large scale geospatial data. Similarly, one can run standard SQL queries on Amazon Athena~\cite{url:athena} or Microsoft Azure for processing spatial data.

\subsection{Future Research Directions}
\label{sub-sec:st_big_future_directions}
Though quite a large number of big data infrastructures have been developed for processing spatial, spatio-temporal, and trajectory data in the last couple of years, most of these systems are not in active development. Also, some of these systems have support for SQL-like queries but not as efficiently as SQL queries in spatial RDBMSs. Besides, their visualization capability is limited except for Python based libraries. Moreover, a few of these systems have support for processing spatial raster data, such as RasDaMan~\cite{Baumann1997_RasDaMan}, SciDB~\cite{Becla2013_SciDB}, GeoTrellis~\cite{url:GeoTrellis}, and Google Earth Engine~\cite{Noel2017_GoogleEE}. Among them, both RasDaMan and SciDB were built from scratch for specific purposes. 

Therefore, considering the volume of spatial raster data generated from various sources (e.g., earth sensors, satellites) and the importance of this data in many application domains, more research is required for processing spatial raster. Also, we need to improve the efficiency of SQL-like queries and the visualization capabilities of existing or new infrastructures. Also, there will be demand in the coming years to include more support for visualizing big spatial data on web platforms. Recently a few big data stream processing platforms (e.g., Apache Flink, Apache Storm) have extended to integrate support for processing spatial or spatio-textual data streams. But more research is required in this area in the coming future. For example, instead of spatial point streams, future research will explore line and polygon streams or will include support for spatio-temporal aspects of data streams. Currently, there is a demand to incorporate SQL-like query engine and dedicated spatial library in DASK. Since the spatial library (cuSpatial) of RAPIDS is in the early stage, there will be more research in the coming future to add more features for processing spatial and spatio-temporal data in GPU. In addition, future big data infrastructures will be more cloud-native and will include machine learning and deep learning models to process spatial data.
\section{Programming and Software Tools for Spatio-temporal Analysis}
\label{sec:st_prog_env}

The infrastructure we use to store spatio-temporal datasets is one of the key aspects of any project involving this type of data. Still, what we do with this data is also of major importance as it allows extracting value out of these data. In this context, it is of utmost importance to review the main programming and software tools that are available to researchers and practitioners for analyzing spatio-temporal data sets.

Analyzing these data may sometimes simply involve
using the available geospatial tools and write some code for storing, querying, analyzing, and visualizing spatial data. Other usage cases involve developing libraries or packages for specific purposes like spatial I/O, visualization, spatial regression, etc. In each of these cases, one essential question comes into our mind: which programming language (or languages) to use to meet our purposes. This decision is frequently driven by our goals. For example, if we want to develop a system that requires heavy-weight development, we will look for a language that is fast and efficient like C/C++ or Java. However, if a developer wants to extend an existing system, the developer will most probably use the language on which the system was built. On the contrary, if our goals involve performing data processing, analysis, and visualization, we need a language that provides a rich set libraries and packages good at these tasks, like Python or R. Moreover, many spatial systems (e.g., ArcGIS) leverage more than one language since some spatial features may be better supported by some languages than others. 

The most popular libraries for modeling spatial data in use today in GIS applications and spatial data processing systems are developed either by Java or C/C++, such as JTS (Java)~\cite{url:JTS}, GEOS (C++)~\cite{url:GEOS}, Google S2 (C++)~\cite{url:GoogleS2}, ESRI Geometry API (Java)~\cite{ESRI_geometry-api}, and Spatial4j (Java)~\cite{url:spatial4j}. Most of the big spatial data processing systems have also utilized these libraries to model spatial data, such as SpatialHadoop (JTS, ESRI Geometry API), GeoSpark (JTS), and GeoMesa (JTS, Spatial4j), etc. Spatial RDBMSs like PostGIS or SpatialLite have used GEOS for modeling spatial data. Even libraries and packages of Python or R have also utilized the GEOS library for modeling spatial data. Recently, researchers have also proposed benchmarks~\cite{Zhang2020_benchmark_gis_libraries, Pandey2020_benchmark_gis_libraries2} for computational geometry libraries used in data processing systems for spatial data exploration.

At present, almost all popular programming languages have some support in terms of libraries or tools~\cite{url:ListGeoLib}, which makes it easier to develop geospatial applications. However, C/C++ and Java are still top in the game for heavy-weight spatial system development. On the other hand, Python or R provide ease of programming and a rich set of libraries and packages for analysis, visualization, and modeling spatial and spatio-temporal data for a wide range of application domains. Also, it is comparatively easy to implement a new library or package for a specific purpose in R and Python, by extending existing support. But if we need to work for the web, then we may need to get support from JavaScript, Python, R, and other web-related languages. It is similar for mobile development; we may need to choose language related to mobile operating systems, such as iOS and Android.


In this section we perform a comprehensive review of widely used libraries and packages of Python and R for analyzing, modeling, and visualizing spatial, spatio-temporal, and trajectory data. We focus on these two programming languages and environments because they are currently seen as the de facto standards for data analysis. This section will also discuss two popular software tools (ArcGIS and QGIS) for spatial data processing. 


\subsection{The R Ecosystem for Spatio-temporal Data Analysis}
\label{sub-sec:st_R}
R is one of the most used languages in data science. From its inception, R is more focused on data analysis and statistical tasks, and therefore, it is more popular with academicians, statisticians, engineers, and scientists, who do not even have prior computer programming knowledge. The capability of R is rapidly growing for statistical analysis, modeling, and visualization of data for a wide range of application domains. In terms of the extension  of libraries and packages for data analysis, R is richer than its counterparts. 

A rich set of packages and libraries are also available in R for analysis and visualization of spatial, temporal, and spatio-temporal data. In addition to analysis and visualization, R also provides interfaces to spatial database systems, GIS software, and big data processing platforms. The R ecosystem for spatio-temporal data analysis is summarized  in Table~\ref{tab:r-st-ecosystem}.

\begin{table}[hbt!]
\centering
\caption{R Ecosystem for Spatio-temporal Data Analysis}
\label{tab:r-st-ecosystem}
\begin{tabular}{|l|l|}
\hline
\textbf{Category} & \textbf{Libraries/Packages/Tools/API's} \\ \hline

Data Processing & 
                \begin{tabular}[c]{@{}l@{}}
                        - sp, sf: spatial vector data \\
                        - raster, terra: spatial raster data \\
                        - spacetime, trajectories, stars: spatio-temporal data \\
                        - xts, zoo, its, ts: time series \\ 
                        - tidygraph: spatial network \\
                        - GDAL (rdal), GEOS (rgeos), PROJ.4(proj4): OSGeo libraries
                \end{tabular} \\ \hline
                
Data Manipulation & - dplyr, tidyr, rmapshaper \\ \hline

Data Modeling & - gstat, CAST, mlr/mlr3, performanceEstimation \\ \hline

Visualization & 
                \begin{tabular}[c]{@{}l@{}}
                        - ggmap: spatial visualization with ggplot2\\
                        - tmap: thematic maps in R (static, animated and interactive)\\
                        - leaflet: JavaScript library for interactive web maps\\
                        - mapview: interactive viewing of spatial data\\
                        - plotly: interactive web-based graphs via plotly.js\\
                        - rasterVis: static visualization of raster data
                \end{tabular} \\ \hline
                
\begin{tabular}[c]{@{}l@{}}APIs for GIS Software\\and Spatial RDBMSs\end{tabular} &
                \begin{tabular}[c]{@{}l@{}}- RQGIS for QGIS\\
                        - RSAGA for SAGA\\
                        - rgrass7 for GRASS\\
                        - RPyGeo for ArcGIS\\
                        - rpostgis for 'PostGIS'-enabled 'PostgreSQL'
                \end{tabular} \\ \hline

\begin{tabular}[c]{@{}l@{}}APIs for\\ Big Data Platforms\end{tabular} &
              \begin{tabular}[c]{@{}l@{}}
                        - Hadoop: Hadoop Streaming, RHadoop, RHIPE, ORCH\\
                        - Spark: SparkR, sparklyr\\
                        - GeoSpark: geospark \\
              \end{tabular} \\ \hline
\end{tabular}
\end{table}

\textbf{Data Processing Infrastructures:} The packages related to spatio-temporal data processing can be categorized into three main groups, namely spatial, temporal, and spatio-temporal. The \textbf{sp}~\cite{Pebesma2005_R_sp} package was the first package related with this type of data developed for R, which consists of methods and classes to represent spatial data types and operations. Since its release in 2005, it became quite popular and nearly 350 other packages are dependent on it. However, spatial features developed in \textbf{sp} are not compliant with OGC simple features~\cite{url:OGC}. Moreover, as many features and functionalities of \textbf{sp} are directly dependent on OSGeo libraries~\cite{url:OSGeo} (such as GDAL, GEOS, and PROJ.4), if these libraries make any changes, it is difficult for \textbf{sp} to manage and maintain interfaces to these libraries due to a lack of simple features. The package \textbf{sf} (simple features)~\cite{Pebesma2018_R_sf} provides classes and methods for spatial vector data, which supersede the \textbf{sp} package. Its features are OGC-compliant and provide direct interfaces to the GDAL, GEOS, and PROJ.4 libraries. Therefore, if we use \textbf{sf},  we do not need to load these external libraries into R code. In addition, \textbf{sf} has many advantages over \textbf{sp}, that include faster I/O operations, improved visualization, compatible with the \textbf{tidyverse} collection of packages (e.g. \textbf{dplyr}), \textbf{sf} objects can be treated as data frames for spatial operations, and finally, the spatial functions in \textbf{sf} have a more consistent naming that makes it easier to use in the code~\cite{lovelace2019_geocomputation_R}. Not surprisingly, \textbf{sf} is quickly being adopted as the backbone for data processing by many other packages related to spatial data analysis.

The \textbf{raster}~\cite{Hijmans2012_R_raster} package is popular for processing spatial raster data. It supports classes and a large set of functions to create, read, write, manipulate, and process raster data. This package can also process large raster datasets that are too large to fit in the main memory. The \textbf{terra}~\cite{Hijmans2020_R_terra} package is a new package for processing raster data in R, containing similar functionality as the \textbf{raster} package. However, due to its simplicity and faster operation,  \textbf{terra} will replace the \textbf{raster} package soon. Besides, this package contains useful methods for data prediction (e.g., interpolation). The packages \textbf{stars}~\cite{Pebesma2020_R_stars} or \textbf{spacetime}~\cite{Pebesma2012_R_spacetime} are used for processing spatio-temporal data, but \textbf{spacetime} also has support for trajectory data. The package \textbf{trajectories}~\cite{Pebesma2020_R_trajectories} was specifically developed for the analysis trajectory data. These packages are  dependent on packages for the analysis of time series (such as \textbf{xts}~\cite{Ryan2020_R_xts}, \textbf{zoo}~\cite{Zeileis2005_R_zoo}) and spatial data (such as \textbf{sp}, \textbf{sf}).

\textbf{Data Manipulation:} The package \textbf{dplyr}~\cite{Wickham2020_R_dplyr} provides a grammar for data manipulation. It contains a set of functions (verbs) to manipulate data in data frames, such as adding new columns, selecting specific columns, filtering rows, re-arranging rows, summarizing data, and other functions. As the \textbf{sf} package is compatible with \textbf{dplyr}, the functions of \textbf{dplyr} can manipulate spatial objects in \textbf{sf}. In addition to manipulating in-memory data frames, \textbf{dplyr} can also manipulate data stored in relational databases (using \textbf{dbplyr}\cite{Wickham2020_R_dbplyr}) or large datasets stored in Spark (using \textbf{sparklyr}~\cite{Luraschi2020_R_sparklyr}).

\textbf{Data Modeling:} The \textbf{gstat}~\cite{Graler2016_R_gstat} package is used for statistical modeling, prediction, and simulation of spatial and spatio-temporal data, which is also dependent on \textbf{sp} package. The \textbf{CAST}~\cite{Meyer2020_R_CAST} package provides functions to improve spatio-temporal modeling tasks using \textbf{caret} package~\cite{Kuhn2008_R_caret}. The \textbf{mlr}~\cite{Bernd2016_R_mlr} package contains a wide range of machine learning algorithms for modeling data. However, due to its complex design, it is difficult for the developers to maintain and add new features in \textbf{mlr}. Besides, as some dependent packages of \textbf{mlr} have changed their features in the meantime, the developers of \textbf{mlr} could not manage to update the \textbf{mlr} accordingly. The \textbf{mlr3}~\cite{Lang2019_R_mlr3} is a successor of \textbf{mlr}, and it is a generic, object-oriented, and extensible framework that solves the above problems. The package \textbf{performanceEstimation}~\cite{Tor2016} also allows for predictive model development and tuning and moreover, contains specific routines for handling predictive tasks with time dependant data, like for instance sliding and growing window model building schemas. 

\textbf{Data Visualization:} Data visualization is an important part of data analysis. In case of spatial data analysis, mapping is one of the best way to present the findings of the research. Though most of the spatial mapping packages were dependent on \textbf{sp} package before, a few of them (such as \textbf{tmap}~\cite{Tennekes2018_R_tmap}, \textbf{leaflet}~\cite{Cheng2019_R_leaflet}, \textbf{mapview}~\cite{Tim2020_R_mapview}) are already supporting the classes of the \textbf{sf} package~\cite{lovelace2019_geocomputation_R}. Spatial maps can be static or interactive and animated. The widely used static mapping tools are \textbf{tmap} and \textbf{ggmap}~\cite{Kahle2013_ggmap}, but \textbf{tmap} is also used for interactive mapping. There is a wide range of packages for interactive and animated maps, such as \textbf{leaflet}, \textbf{mapview} and \textbf{plotly}~\cite{Sievert2020_R_plotly}. Besides, \textbf{rasterVis}~\cite{Oscar2020_R_rasterVis} is a common package for static raster data visualization.    

\textbf{APIs for GIS Software and Spatial RDBMSs:} R is very rich in terms of libraries and packages for analysis spatio-temporal data, but it is neither a spatial database system nor a powerful standalone GIS software tool. Besides, R packages are not capable of processing large spatio-temporal data. Therefore, the integration of R with GIS software and Spatial RDBMSs extends the capabilities of R for processing spatio-temporal data. As a result, R users can use hundreds of algorithms from GIS software and can process data stored in database systems using a rich set of R packages. \textbf{RQGIS}~\cite{Muenchow2017_RQGIS} establishes an interface between R and QGIS~\cite{QGIS_software} by utilizing Python API for QGIS. It provides access to QGIS algorithms from within R. As QGIS has already integrated other popular GIS software (such as GDAL, SAGA, GRASS, and more), integrating R with QGIS brings the power of all these software into R using only one package, \textbf{RQGIS}. However, one can use dedicated APIs to access complete support of each of these GIS software that include \textbf{rgdal}~\cite{Bivand2020_rgdal} for GDAL, \textbf{RSAGA}~\cite{Brenning2018_RSAGA} for SAGA, and \textbf{rgrass7}~\cite{Bivand2019_rgrass7} for GRASS. R users can also access functionalities of another popular commercial GIS software ArcGIS through the \textbf{RPyGeo}~\cite{Brenning2018_RPyGeo} package. Spatial RDBMSs (e.g., PostgreSQL/PostGIS, Oracle Spatial) can store, manage, and query both vector and raster data efficiently. However, their data analysis and visualization capabilities are very limited. Therefore, the integration of R with spatial RDBMSs is useful for both R and spatial RDBMSs users. The \textbf{rpostgis}~\cite{Bucklin2018_rpostgis} package provides R with an interface to access a popular open-source database system, PostgreSQL/PostGIS. It also supports methods to perform read and write operations with PostgreSQL/PostGIS for handling both vector and raster spatial datasets.

\textbf{APIs for Big Data Platforms:} Traditional GIS software and spatial RDBMSs are not capable of handling today's huge volume of multi-dimensional and heterogeneous spatio-temporal data. Therefore, R interfaces to these systems do not provide scalability and efficiency for processing big spatio-temporal data. There are a number of R APIs that provide access to scalable and fault-tolerant distributed big data computing platforms, such as Hadoop and Spark. We can use \textbf{Hadoop Streaming}~\cite{url:HStream}, \textbf{RHadoop}~\cite{url:RHadoop}, \textbf{RHIPE} (R and Hadoop Integrated Programming Environment)~\cite{url:RHIPE}, and \textbf{ORCH} (Oracle R Connector for Hadoop)~\cite{url:ORCH} API to run MapReduce jobs using Hadoop or directly accessing Hadoop Distributed file System (HDFS) within the R programming environment. However, the Hadoop Streaming API can  only run R MapReduce scripts that are written using the  \textbf{HadoopStreaming}~\cite{Rosenberg2012_HadoopStream} package. The  \textbf{sparklyr}~\cite{Luraschi2020_R_sparklyr} and \textbf{SparkR}~\cite{Shivaram2016_SparkR} packages are used as an interface between R and Spark. The package \textbf{sparklyr} is compatible with \textbf{dplyr} and allows R users to access the built-in machine learning algorithms of Spark. \textbf{SparkR} is a native Spark frontend for R users that provide access to all Spark libraries. Besides, these packages also allow R users to access HDFS. However, both Hadoop and Spark do not have native support for processing spatial data. On the other hand, most R packages for spatial analysis are developed for a single node. Therefore, we can implement a custom R package for Hadoop and Spark-based spatial data processing systems (discussed in Section~\ref{sub-sec:hadoop-systems} and~\ref{sub-sec:spark-systems}) to achieve scalability for spatial processing. For example, \textbf{geospark}~\cite{url:rgeospark} package allows R users to use GeoSpark~\cite{Yu2015_GeoSpark1} for big spatial analysis.


\subsection{Python Ecosystem for Spatio-temporal Data Analysis}
\label{sub-sec:st_python}
Python is a full-fledged general purpose programming language. However, Python has become one of the most popular programming languages for data science in the last decade. It is widely used for processing, analyzing, and visualizing data in both academia and industry. Researchers and organizations are continuously working to develop new tools, libraries, and packages to process and analyze real-world data. Therefore, the community of Python is growing as well as the capabilities of the language. Other than data science, Python is also popular with programmers and developers for developing general-purpose software applications. Due to the rise of spatial data, Python has been adopted for modeling, analyzing, and visualizing spatial, temporal, and spatio-temporal data in the last decade. The integration of Python as the main scripting language by popular GIS platforms like ArcGIS and QGIS has expedited this process. Python also supports interfaces to big data computing platforms, Hadoop and Spark. The Python ecosystem for spatio-temporal data analysis is summarized in Table~\ref{tab:python-st-ecosystem}.

\begin{table}[hbt!]
\centering
\caption{Python Ecosystem for Spatio-temporal Data Analysis}
\label{tab:python-st-ecosystem}
\begin{tabular}{|l|l|}
\hline

\textbf{Category} & \textbf{Libraries/Packages/Tools/API's} \\ \hline

Core Libraries &
        \begin{tabular}[c]{@{}l@{}}
                - SciPy: core library for scientific computation\\ 
                - NumPy: fundamental library for numerical computation\\ 
                - Pandas: data structure and analysis library
        \end{tabular} \\ \hline

Data I/O &
        \begin{tabular}[c]{@{}l@{}}
                - GDAL: raster and vector I/O (interface to GDAL/OGR)\\ 
                - Fiona: vector I/O (interface to OGR)\\ 
                - Rasterio: raster I/O (interface to GDAL)
        \end{tabular} \\ \hline
        
Data Processing &
        \begin{tabular}[c]{@{}l@{}}
                - GeoPandas: spatial extension of pandas \\ 
                - Shapely: spatial analysis of geometric objects\\ 
                - scipy.spatial: spatial algorithms and data structures\\ 
                - pyspatial: analysis vector/raster data\\ 
                - sptemp: spatio-temporal vector data analysis\\ 
                - Rtree: spatial indexing\\ 
                - rasterstats: summarizing spatial raster datasets\\ 
                - MovingPandas, traja: trajectory data analysis\\
                - pyproj: coordinate transformations (interface to PROJ4)
        \end{tabular} \\ \hline
        
\begin{tabular}[c]{@{}l@{}}Statistical Analysis\\and Modeling\end{tabular} &
        \begin{tabular}[c]{@{}l@{}}
                - PySAL: spatial and spatio-temporal data analysis library\\ 
                - scikit-learn: machine-learning algorithms\\ 
                - scikit-image: algorithms for image (satellite) processing\\ 
                - statsmodels: statistical modeling for python
        \end{tabular} \\ \hline
        
Visualization &
        \begin{tabular}[c]{@{}l@{}}
                - Matplotlib: static and interactive visualization\\
                - Seaborn: statistical data visualization\\ 
                - Bokeh: interactive visualizations for the web\\ 
                - Plotly: interactive visualizations for the web\\ 
                - Folium: visualizations via interactive leaflet map (leaflet.j)\\ - Cartopy: visualize data on maps\\ 
                - ggplot: visualizations based on R ggplot2
        \end{tabular} \\ \hline
        
\begin{tabular}[c]{@{}l@{}}APIs for\\GIS Software\end{tabular} &
        \begin{tabular}[c]{@{}l@{}}
                - ArcPy, ArcGIS API: python interface to ArcGIS\\ 
                - PyQGIS: python interface to QGIS
        \end{tabular} \\ \hline
        
\begin{tabular}[c]{@{}l@{}}APIs for\\Big Data Platforms\end{tabular} &
        \begin{tabular}[c]{@{}l@{}}
                - Hadoop: Hadoop Streaming, mrjob, Pydoop, Luigi, PyArrow\\
                - Spark: PySpark
        \end{tabular} \\ \hline
\end{tabular}
\end{table}

The SciPy stack~\cite{Pauli2020_SciPy} consists of a set of libraries for scientific computing in Python. Specifically, the  \textbf{SciPy}~\cite{Pauli2020_SciPy}, \textbf{Numpy}~\cite{2020NumPy-Array}, and \textbf{Pandas}~\cite{reback2020_Pandas, mckinney2010_Pandas} libraries have been used as core packages in data science. These packages are also  essential for the analysis of spatio-temporal data as most of the libraries of the Python spatial stack depend on them.

\textbf{Data I/O:} The spatial input/output libraries of Python are developed using the existing C library, GDAL (Geospatial Data Abstraction Library)~\cite{Team2020_GDAL}, which supports a wide range of raster and vector data formats. Therefore, Python spatial I/O libraries also support these data formats. \textbf{Fiona}~\cite{Gillies_fiona} interfaces to the  OGR (OpenGIS Reference Implementation) layer of GDAL for reading and writing spatial vector data of various formats, such as Shapefile, GeoJSON, JSON, CSV, etc. The library  \textbf{rasterio}~\cite{Team2018_rasterio} interfaces to GDAL for raster functionality. It relies on \textbf{Numpy} for efficient processing of raster formats, such as GeoTIFF, netCDF, JPEG2000, and other formats. 

\textbf{Data Processing:} A set of libraries is available in Python for processing spatial and spatio-temporal data. The library \textbf{Shapely}~\cite{Team_shapely} provides functions for manipulation and analysis of vector geometric objects, and is based on the widely used GEOS~\cite{url:GEOS} library. \textbf{GeoPandas}~\cite{Team_GeoPandas} is a spatial extension of Pandas. It uses the Shapely, Fiona, and pyproj (Python interface to PROJ.4)~\cite{Team_pyproj} libraries to add spatial support in the popular data analysis and manipulation tool, Pandas. \textbf{SciPy.Spatial} also provides algorithms and data structures for spatial analysis. Whereas, \textbf{rasterstats}~\cite{Perry2015_rasterstats} contains functions for zonal statistics and interpolated point queries for summarizing spatial raster datasets using vector geometries. It can work with any data formats supported by rasterio. Moreover, one can use \textbf{pyspatial}~\cite{Aman2016_pyspatial} for both raster and vector data, \textbf{sptemp}~\cite{url:sptemp} for spatio-temporal vector data, and \textbf{MovingPandas}~\cite{Anita2019_MovingPandas} and \textbf{traja}~\cite{Justin_2019_traja} for trajectory data analysis.

\textbf{Statistical Analysis and Modeling:} 
\textbf{PySAL}~\cite{Rey2007_pysal} is an open-source spatial analysis library with a primary focus on vector data. The functionality of PySAL covers a wide range of areas, such as methods to detect spatial clusters, hot spots and outliers, spatial regression, statistical modeling, spatial econometrics, space-time analysis, visualization, and more. The current version of PySAL (v2.X) consists of four domains, which include PySAL core (pysal.lib), exploratory spatial data analysis (pysal.explore), spatial statistical models (pysal.model), and geovisualization (pysal.viz). \textbf{pysal.lib} is the core library which contains data structures and algorithms for spatial I/O, spatial weights, computational geometry, and more. The \textbf{pysal.explore} library consists of modules for exploratory analysis of spatial and spatio-temporal data. The \textbf{pysal.model} is designed to model spatial relationships in data using different types of linear, generalized-linear, generalized-additive, nonlinear, multi-level, and local regression models. Finally, the \textbf{pysal.viz} layer supports functionality to visualize spatially analyzed data (e.g., detected clusters or hot-spots). Besides, PySAL provides a toolkit for ArcGIS and a plugin for QGIS which allows using the functionalities of PySAL within these GIS software. Some desktop applications like CAST (Crime Analytics in Space-Time) and GeoDaSpace has also use a subset of PySAL. Moreover, PySAL is now available as a featured package in the distribution of Anaconda Python and Enthought Canopy~\cite{Rey2015_OGA_PySAL}. \textbf{statemodels}~\cite{seabold2010_statsmodels} is another useful library for users who are looking for a Python library for statistics, financial econometrics, or econometrics. It also supports models for time-series analysis.

\textbf{scikit-learn}~\cite{scikit-learn_2011} is a library of a vast collection of supervised and unsupervised machine learning algorithms for clustering, classifications, regression, dimensionality reductions, and many more. It also supports functions for data loading, manipulation, and prepossessing. Whereas \textbf{scikit-image}~\cite{scikit-image_2014} includes a wide range of algorithms for image analysis, such as segmentation, transformations, restoration, metrics, feature selection, color space manipulation, filtering, morphology, and other algorithms. Since both of these libraries are implemented based on the libraries of the SciPy ecosystem, SciPy libraries must be installed before using scikit-learn and scikit-image.

\textbf{Data Visualization:} Like R, Python also supports a rich set of libraries for spatial data visualization and mapping. \textbf{Matplotlib}~\cite{Hunter2007_Matplotlib} is a key visualization library in Python (part of a SciPy ecosystem), which can be used for creating static, interactive, and animated visualizations. The package \textbf{Cartopy}~\cite{Cartopy} uses PROJ.4, Shapely, and Numpy to provide a spatial mapping library on top of Matplotlib. It includes an easy way to create maps via Matplotlib. \textbf{Seaborn}~\cite{michael2017_seaborn} is another library built on top of Matplotlib for statistical data visualization. A number of Python packages are also built to create interactive maps for the web, such as Folium~\cite{Rob_folium}, Bokeh~\cite{Team2020_Bokeh}, and Plotly~\cite{Sievert2020_R_plotly}. \textbf{Folium} is a wrapper for the \textbf{leaflet.js} library for plotting interactive web maps. It includes a raster and a vector layer for visualizing through an interactive leaflet map. Similarly, \textbf{Bokeh} and \textbf{Plotly} are also developed as interactive visualization libraries for web browsers. 


\textbf{Python for GIS Software:} The integration of Python as a main scripting language of \textbf{ArcGIS}~\cite{url:ArcGIS} and \textbf{QGIS}~\cite{QGIS_software} allows users to use the combined power of GIS software and Python for processing spatial data. The \textbf{ArcPy} package ships with a desktop version of ArcGIS and allows Python to access GIS tools with extensions, useful functions, classes, and modules for processing geospatial data. This package helps users to write Python scripts which can be run within ArcGIS or as  standalone scripts.  \textbf{ArcGIS API} for Python (called Pythonic GIS API) gives access to a wide range of modules, classes, and functions provided by ArcGIS Online and ArcGIS Enterprise for web-based GIS solutions. Similarly, we can use Python with QGIS in many ways, such as executing commands in the Python console within QGIS, extending the functionality of QGIS by developing new plugins, creating standalone Python scripts, and developing custom GIS applications using \textbf{PyQGIS API}~\cite{url:PyQGIS}. Other popular GIS software like \textbf{GRASS GIS}~\cite{url:GRASS} and \textbf{SAGA GIS}~\cite{url:SAGA} also support APIs for Python.

\textbf{APIs for Big Data Platforms:} As it was mentioned in Section~\ref{sec:intro}, spatial libraries and packages of Python were developed for processing data in a single-node environment. Therefore, we need to use  parallel and distributed computing platforms like Hadoop and Spark for processing larger sets of data. There are a number of Hadoop APIs that allow Python users to access the Hadoop MapReduce paradigm and distributed file system HDFS, which include \textbf{Hadoop Streaming}, \textbf{mrjob} (Yelp)~\cite{Team2020_mrjob}, \textbf{Pydoop}~\cite{Leo2010_Pydoop}, and \textbf{Luigi} (Spotify)~\cite{Team2020_Luigi}. One can write and run MapReduce jobs on Hadoop using all these APIs, but the Hadoop Streaming API ships with Hadoop as a native API. \textbf{mrjob} can also run MapReduce jobs locally without Hadoop for testing purposes.  \textbf{Pydoop} is tightly integrated with Hadoop and provides full access to Hadoop APIs. Pydoop also supports direct access to HDFS via its HDFS API. Moreover, \textbf{PyArrow} also includes an HDFS client to access HDFS. On the other hand, \textbf{PySpark}~\cite{Team2020_PySpark} is a native Spark API that enables Python users to interact with the Spark programming paradigm for processing large datasets. Along with a rich set of Python libraries, this API also allows Python users to use built-in Spark libraries, such as MLlib (machine learning), Spark Streaming, and Spark SQL and Dataframes. Besides, PySpark can process data stored in a distributed storage like HDFS. However, these Python APIs add an extra overhead in computation during big spatial data processing as they are developed as a layer on top of Hadoop and Spark. Hence, \textbf{DASK}~\cite{Team2016_Dask} and \textbf{RAPIDS}~\cite{2018_RAPIDS} have emerged as parallel and distributed Python libraries to mitigate this issue (see Section \ref{sub-sec:python_big_library}).


\subsection{GIS Software}
\label{sub-sec:st_gis_softwares}

A GIS (Geographical Information System) is an integrated environment to capture, store, analyze, and visualize all kinds of geographical data (raster, vector, and network). GIS software is a useful tool for researchers, scientists, or practitioners, who want to extract inherent knowledge, patterns, and relationships from geographical data and analyze the data to address real-world problems. There are a few commercial and open-source GIS software applications available, such as ArcGIS~\cite{url:ArcGIS}, QGIS~\cite{QGIS_software}, GRASS~\cite{url:GRASS}, and SAGA~\cite{url:SAGA}. ArcGIS and QGIS are the most popular GIS software among them.

\textbf{ArcGIS}~\cite{url:ArcGIS} is a leading commercial GIS software application developed by ESRI~\cite{url:esri}. Whereas, \textbf{QGIS} (Quantum GIS)~\cite{QGIS_software} is a popular open-source GIS software application that supports similar functionalities to ArcGIS. The desktop version of ArcGIS only supports the Windows platform, but QGIS is available for all popular computing platforms like Windows, Mac, and Linux. As QGIS has integrated a few other popular GIS software like GRASS~\cite{url:GRASS}, SAGA~\cite{url:SAGA}, and OTB (Orfeo Toolbox)~\cite{Grizonnet2017_Orfeo}, one can use a subset of algorithms from these third party GIS software within QGIS. However, if we want to use full functionality of these GIS software systems, we need to stick with GRASS and SAGA~\cite{lovelace2019_geocomputation_R}. Both ArcGIS and QGIS have support for a Python console that allows users to execute the functionality of GIS tools and Python within ArcGIS and QGIS. Also, both of them have an interface to Python and R. Therefore, one can use the functionality of ArcGIS and QGIS within Python and R programming environments. According to an experiment conducted by Debicka et al.~\cite{Justyna2018_ArcGISvsQGIS} based on the buffer, convex hull, and intersection geometric operations, QGIS is faster than ArcGIS for processing spatial data. However, in terms of spatial mapping capabilities, ArcGIS is way better than QGIS. ArcGIS is also richer in terms of tools and supported algorithms, but one can extend the capabilities of QGIS by adding third-party plugins. Moreover, the supported input file format of QGIS is very rich, as it uses the GDAL/OGR library.

Like Spatial RDBMSs, GIS software is also going through many changes over the years to adapt to this era of big spatial data. ESRI has released open-source GIS tools for Hadoop~\cite{url:gis-tools-for-hadoop} to perform analysis on big spatial data by utilizing the distributed processing capability of Hadoop. These tools include: (i) a geometry API for Java users to develop MapReduce applications for spatial data analysis; (ii) a spatial framework, which allows users to perform SQL-like queries on spatial data using the Hive Query Language (HQL); (iii) a geoprocessing toolbox, which allows users to take advantage of both Hadoop and ArcGIS for spatial processing; and (iv) a complete toolkit containing a geometry API, a spatial framework, and a  geoprocessing toolbox. ESRI has also introduced a Spark-powered GeoAnalytics toolbox for both ArcGIS server and desktop versions. GeoAnalytics Desktop~\cite{url:GeoAnalytics-Desktop} brings parallel processing of data across multiple cores of a personal computer through ArcGIS Pro. Whereas, GeoAnalytics Server~\cite{url:GeoAnalytics-Server} provides distributed processing of big spatial data across multiple nodes of a cluster running ArcGIS Enterprise. As Spark ships with ArcGIS, users do not need to install Spark separately. Users can also write a Python program to process data using the GeoAnalytics toolbox. Besides, ESRI ArcSDE is an RDBMSs gateway that allows ArcGIS users to store, use, and manage spatial data in some popular databases, such as IBM DB2 and Informix, Oracle, Microsoft SQL Server, and PostgreSQL. In summary,  GIS software is continuously adapting to process big spatial data. 

\subsection{Future Research Directions}
\label{sub-sec:st_prog_future_research}

At present, Python and R support a rich set of libraries and packages for processing spatio-temporal data. These libraries and packages are developed for processing data in a single node computer system and are not suitable for processing big data. Python and R users could utilize the available APIs for big data processing platforms, such as Hadoop and Spark. But we do not know how these libraries will perform with big data platforms since there is no comprehensive evaluation yet. Besides, these APIs will add an extra overhead in overall the computation since we need to compile Python or R code into the target platform. We can also implement APIs to use big spatio-temporal systems that have been developed based on big data platforms. Currently, DASK and RADPIS are promising platforms to process big spatial data for Python users. R users either need to use an existing big spatio-temporal data processing system or need to develop a system like DASK.

On the other hand, GIS software will be adding new features and modules to create analysis and mapping facilities for a wide range of new application domains. As machine learning (ML) and deep learning (DL) algorithms and techniques are important for solving complex spatial problems, there will be more ML and DL models in GIS software in the future. Therefore, along with parallel and distributed computing, more research is required in terms of integrating ML and DL tools with GIS Software.
\section{Conclusion}
\label{sec:conclusion}
Due to the rise of spatio-temporal data volume and the significance of extracted knowledge in a wide range of application domains, plenty of research and development works have been done in the area of spatio-temporal data analytics in the past decade. Survey work is always pivotal for researchers to know and advance the state of the art. In this survey, we have conducted a comprehensive study on the whole ecosystem of spatio-temporal data analytics, which covers spatial databases (SQL and NoSQL), big spatial-temporal infrastructures, programming languages, and software tools. This study also addressed the importance, current demand, and future of spatio-temporal data analytics. Though the main focus was on spatio-temporal analysis for big data, we have discussed  related areas as well. 

We argue that the research community needs to address a few areas of spatio-temporal data analytics in future research that include (i) integrating more support to model and analysis of spatial raster data, (ii) integrating more support for processing spatio-temporal (trajectory) data streams, (iii) integrating or improving SQL-like queries, and (iv) adding more support for analysis and visualization in big spatio-temporal infrastructures. It is already evident that there will be more research on integrating AI, machine learning, and deep learning models in future big spatio-temporal infrastructures for uncovering hidden knowledge. There will be demand for integrating visualization support for big spatial data in web platforms. Besides, we think future infrastructures will be more application-specific, such as IoT, neuroscience, emergency management, transportation, and other applications. The usage of GPU in RAPIDS has shown significant speed-up in computation compared to other infrastructures, and therefore, more research is required in terms of using GPUs for spatio-temporal data analytics. We hope that the accumulated information in this study will be useful for researchers, practitioners, and developers who are currently working or who want to work in the area of spatio-temporal data analytics.

\begin{acks}
The work of L. Torgo was undertaken, in part, thanks to funding from the Canada Research Chairs program and NSERC.
\end{acks}

\bibliographystyle{ACM-Reference-Format}
\bibliography{acm-main}


\end{document}